\documentclass{article}

\usepackage{microtype}
\usepackage{graphicx}
\usepackage{subcaption}
\usepackage{booktabs} %

\usepackage{hyperref}

\usepackage[preprint]{icml2026}

\usepackage{amsmath}
\usepackage{amssymb}
\usepackage{mathtools}
\usepackage{amsthm}

\usepackage[capitalize,noabbrev]{cleveref}

\usepackage{amsfonts}
\usepackage[linesnumbered,ruled,vlined]{algorithm2e}
\usepackage[inline]{enumitem}   %
\usepackage{booktabs}
\usepackage{adjustbox}
\setlist{nosep}
\usepackage{tcolorbox}
\usepackage{subcaption}
\usepackage[dvipsnames]{xcolor}
\usepackage{wrapfig}
\usepackage{dsfont}

\definecolor{mypurple}{RGB}{153, 0, 255}

\crefname{algocf}{alg.}{algs.}
\Crefname{algocf}{Algorithm}{Algorithms}

\usepackage{amsmath,amsfonts,bm}

\def\eqref#1{equation~\ref{#1}}

\def\1{\bm{1}}

\DeclareMathAlphabet{\mathsfit}{\encodingdefault}{\sfdefault}{m}{sl}
\SetMathAlphabet{\mathsfit}{bold}{\encodingdefault}{\sfdefault}{bx}{n}

\newcommand{\E}{\mathbb{E}}

\newcommand{\R}{\mathbb{R}}

\newcommand{\Var}{\mathrm{Var}}

\newcommand{\Cov}{\mathrm{Cov}}

\theoremstyle{plain}
\newtheorem{theorem}{Theorem}[section]

\newtheorem{lemma}[theorem]{Lemma}
\newtheorem{corollary}[theorem]{Corollary}
\theoremstyle{definition}

\theoremstyle{remark}

\usepackage[textsize=tiny]{todonotes}

\icmltitlerunning{Uncertainty Estimation via Transformation Equivariance}

\begin{document}

\twocolumn[
  \icmltitle{Uncertainty Estimation for Pretrained Medical Image Registration Models \\ via Transformation Equivariance}

  \icmlsetsymbol{equal}{*}

  \begin{icmlauthorlist}
    \icmlauthor{Lin Tian}{1}
    \icmlauthor{Xiaoling Hu}{1}
    \icmlauthor{Juan Eugenio Iglesias}{1,2,3}
  \end{icmlauthorlist}

  \icmlaffiliation{1}{Massachusetts General Hospital and Harvard Medical School}
  \icmlaffiliation{2}{The Hawkes Institute, University College London}
  \icmlaffiliation{3}{Computer Science and AI Laboratory, Massachusetts Institute of Technology}

  \icmlcorrespondingauthor{Lin Tian}{ltian3@mgh.harvard.edu}

  \icmlkeywords{Uncertainty Estimation, Medical Image Registration}

  \vskip 0.3in
]

\printAffiliationsAndNotice{}  %

\begin{abstract}
Accurate image registration is essential in many medical imaging applications, yet most deep registration networks provide little indication of when or where their predictions are unreliable. Existing uncertainty estimation approaches, such as Bayesian methods, ensembles, or MC-dropout, typically require architectural modifications or retraining, precluding their applicability to pretrained registration models. We propose an inference-time, model-agnostic uncertainty estimation framework that applies directly to \emph{any} pretrained registration network. Our approach is grounded in the transformation equivariance property of image registration, which states that the underlying anatomical mapping should remain consistent under spatial perturbations of the input. Experiments across three pretrained registration models and four anatomical structures show that the resulting uncertainty maps consistently correlate with registration error and highlight unreliably aligned regions. This framework turns pretrained registration networks into risk-aware tools at test time, moving medical image registration closer to safe clinical and large-scale research deployment.

\end{abstract}

\section{Introduction}
Accurate alignment of images (image registration) is a cornerstone of contemporary medical imaging pipelines. It enables change detection between longitudinal scans for disease monitoring \citep{leung2010automated,tustison2019longitudinal}, links patient anatomy to population atlases \citep{dubost2020multi}, tracks lesions across longitudinal scans \citep{rokuss2025lesionlocator}, propagates dose maps for radiation \citep{kessler2006image}, and underpins nearly every quantitative morphometric analysis.  Modern learning-based approaches can often achieve real-time registration with performance comparable to, and in some benchmark settings exceeding, that of classical methods~\citep{hering2022learn2reg,chen2025beyond}. However, most of them remain \emph{opaque}: they predict a deformation field from an image pair, but \emph{provide little indication of where or when that prediction may be unreliable}. Visual quality control of registrations is extremely labor intensive (particularly in 3D) and sometimes unfeasible (e.g., in real-time applications). Therefore, silent misregistrations can corrupt downstream measurements and go unnoticed until late in clinical or research workflows. In safety-critical settings, identifying regions where predictions are unreliable is as important as producing the registration itself.

In the deep learning era, uncertainty estimation for image registration has primarily been explored through Bayesian deep learning techniques, such as variational inference, Monte Carlo dropout, or ensembles over network parameters~\citep{dalca2019unsupervised,sedghi2019probabilistic,yang2017quicksilver,gong2022uncertainty,smolders2022deformable}. While effective, these approaches typically require retraining the model, modifying the network architecture, or accessing the original training data. Such requirements make them impractical for pretrained registration models that are deployed as fixed components in downstream pipelines. Moreover, retraining deployed models may invalidate retrospective studies that rely on consistent processing across cohorts. These challenges call for uncertainty estimation methods that operate at inference time, without altering pretrained models. Furthermore, it remains unclear how the estimated uncertainties produced by existing methods relate to actual registration errors~\citep{luo2019applicability,luo2020registration}.

\begin{figure}
    \centering
    \includegraphics[width=0.85\linewidth]{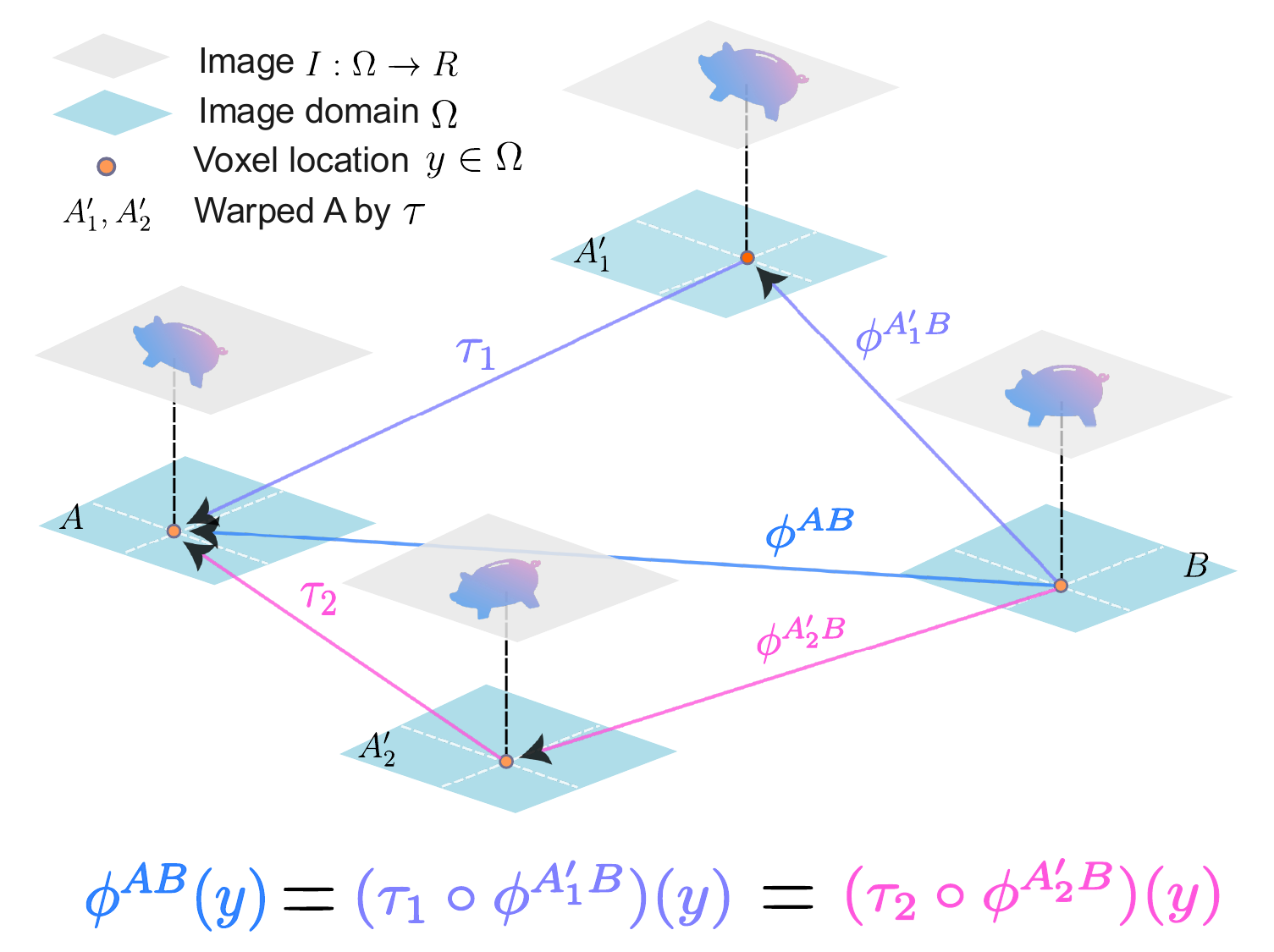}
    \caption{Transformation equivariance for image registration. The mapping between image A and B should stay consistent regardless of whether we warp image A by $\tau$ or not.}
    \label{fig:transformation_equivariance}
    \vspace{-.2in}
\end{figure}

To address these limitations, we propose an inference-time, model-agnostic uncertainty estimation framework that applies directly to \emph{any} pretrained image registration model. Our approach is grounded in the principle of \emph{transformation equivariance} for image registration, which states that a registration mapping should behave consistently when the input is spatially perturbed (\cref{fig:transformation_equivariance}). Intuitively, if we perturb a source image before registration, a robust model should produce a deformation field that, once the initial perturbation is factored out, remains consistent with its original prediction. Building on this principle, by measuring the variability of these corrected predictions across a distribution of different spatial perturbations, we can quantify the model's uncertainty in finding the geometric correspondence between equivalent image pairs. Crucially, the proposed framework requires \textbf{no retraining or architectural modifications}, making it immediately applicable to pretrained registration models, including those not equipped with probabilistic formulations.

\begin{figure*}[t]
    \centering
    \includegraphics[width=\linewidth]{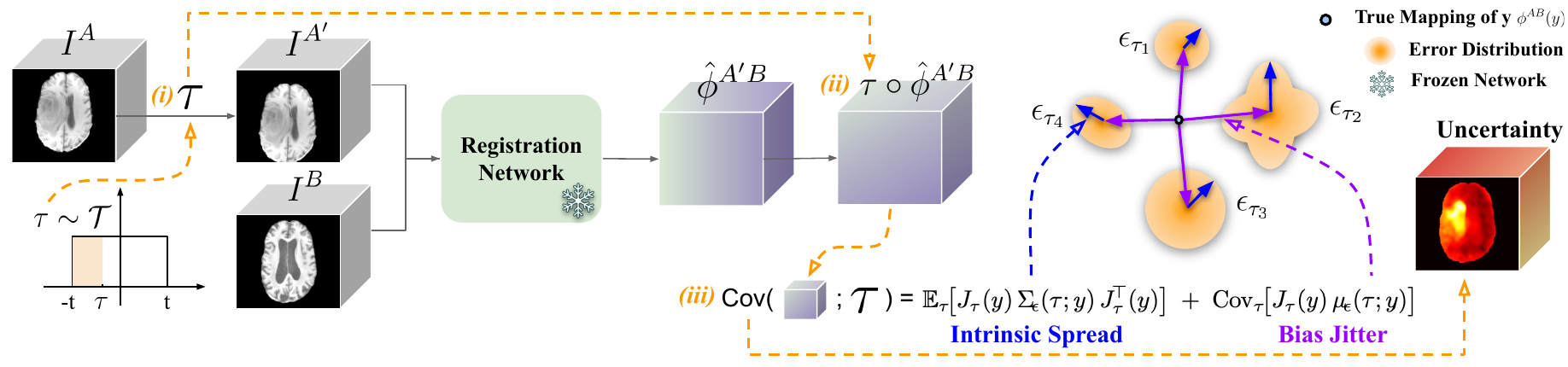}
    \caption{Overview of the proposed uncertainty estimation framework. We present an inference-time, model-agnostic framework for \textcolor{orange}{uncertainty estimation} in image registration that requires no model retraining or architectural modifications. Concretely, we \textcolor{orange}{\textit{(i)}} perturb the source image coordinate frame $I^{A^{\prime}}=I^A\circ\tau$, \textcolor{orange}{\textit{(ii)}} compose the predicted transformations $\tau\circ\hat{\phi}^{A^{\prime}B}$ to represent the mapping between the original images $I^A$ and $I^B$, and \textcolor{orange}{\textit{(iii)}} measure their variance $\operatorname{Cov}_\tau[\tau\circ\hat{\phi}^{A^{\prime}B}]$ to produce uncertainty maps. This variance decomposes into two interpretable components: \textcolor{blue}{Intrinsic Spread}, reflecting the average local variability of the error distribution, and \textcolor{mypurple}{Bias Jitter}, capturing the shifts of the error distributions under spatial perturbations.}
    \label{fig:teaser}
\end{figure*}

In summary, our main contributions are:  
\begin{itemize}
    \item We propose an inference-time uncertainty estimation framework that applies to arbitrary pre-trained image registration models without retraining (\cref{sec:method}).  
    \item We provide a theoretical analysis of the proposed uncertainty map, deriving a decomposition of the variance that characterizes its relationship with registration error (\cref{sec:theory_analysis}). 
    \item We validate the proposed uncertainty estimation framework through extensive experiments on four anatomical structures (brain, cardiac, abdominal, and lung) with three pre-trained registration models (uniGradICON, SynthMorph, TransMorph), demonstrating (\emph{i}) strong empirical agreement with MC-dropout uncertainty (\cref{sec:compare_with_mcdropout}),  (\emph{ii}) strong correlation between estimated uncertainty and true registration error (\cref{sec:dataset_level_evaluation}), and (\emph{iii}) informative behavior in the presence of anatomical inconsistencies (\cref{sec:exp_case_study}).
\end{itemize}

Together, these results suggest that the proposed uncertainty estimation provides a simple, model-agnostic tool to probe the reliability of registration networks at inference time, moving dense registration one step closer to safe deployment in clinical and large-scale research settings. 

\vspace{-.05in}
\section{Related Work}
\textbf{Learning-based medical image registration.}
Medical image registration aims to estimate a spatial transformation that aligns two given images. Classical registration methods \citep{avants2008symmetric,klein2009elastix,modat2010fast,heinrich2012globally} typically formulate this as an optimization problem over a set of parameters describing the transformation. They can be used for a wide variety of registration tasks and can be highly accurate, but are often slow as they estimate registration parameters from scratch for every registration pair by numerical optimization. More recent supervised \citep{yang2017quicksilver,cao2017deformable,sokooti2017nonrigid} and unsupervised \citep{de2017end,balakrishnan2019voxelmorph} learning-based registration approaches \emph{predict} spatial correspondences much faster using a deep registration network. These learning-based approaches have achieved significant accuracy improvements by advanced transformation models \citep{shen2019networks,niethammer2019metric, tian2024nephi}, network architectures \citep{mok2020fast,chen2022transmorph}, training schemes \citep{hering2019mlvirnet,de2019deep,shen2019networks,mok2020large}, similarity measures \citep{tian2023same++,mok2024modality,song2024dino}, and regularization \citep{greer2021icon, tian2023gradicon}. However, compared to the volume of work on learning-based registration networks, few works have explored uncertainty estimation in medical image registration, as we discuss in the following paragraphs. 

\textbf{Uncertainty estimation in deep learning.}
Uncertainty estimation aims to estimate how confident a neural network is in its predictions, serving as an important tool when deploying networks to the real world. Classical Bayesian approaches approximate the posterior distribution of model parameters via variational inference or stochastic regularization such as MC-dropout \citep{gal2015bayesian}. Deep ensembles provide another powerful non-Bayesian alternative \citep{lakshminarayanan2017simple, rupprecht2017learning}. Consistency \citep{moon2020confidence, li2023confidence} has also been used to measure uncertainty. These techniques offer valuable insights, but typically require architectural changes, retraining (and possibly access to the original data), or significant computational cost, making them difficult to apply to pretrained models. 

\textbf{Uncertainty estimation with test-time augmentation (TTA).}
As an alternative, TTA estimates uncertainty by measuring prediction variability under input perturbations \citep{ayhan2018test, wang2019aleatoric}. In tasks like classification or segmentation, this variability primarily reflects a model’s sensitivity to data noise, such as spatial or appearance variance. However, in registration, the network predicts a spatial transformation itself. Consequently, variance under spatial perturbation reveals more than just data sensitivity; it exposes the fundamental limitations of the model's \emph{geometric reasoning}, which is shaped by its network architecture, similarity measures, and regularization priors. Thus, while conceptually related to TTA, the proposed framework captures uncertainty in anatomical correspondence across equivalent image pairs. This uncertainty arises from factors such as inherent image ambiguity, the model's inability to extract representative features for stable correspondence, or restrictive regularization priors for large-magnitude deformations, limitations that cannot be explained away simply by increasing training-time augmentation.

\textbf{Uncertainty estimation for medical image registration.}
Uncertainty estimation for image registration has long been an object of study. In the classical literature, probabilistic modeling approaches estimate uncertainty by computing (exactly or approximately) posterior distributions over deformation parameters via Bayesian inference \citep{simpson2013bayesian, kybic2009bootstrap, le2017sparse, risholm2013bayesian, le2016quantifying, agn2019fast} or Monte Carlo sampling \citep{iglesias2013improved}.
Alternatively, bootstrap sampling has been employed as an empirical ensemble strategy to quantify uncertainty \citep{kybic2009bootstrap}.

In the deep learning era, however, uncertainty estimation remains largely underexplored. Most approaches either directly adapt classical techniques, e.g., Bayesian inference or ensembles \citep{yang2017quicksilver,gong2022uncertainty, smolders2022deformable, chen2024registration} (see also the survey by \citealt{chen2024surveydeeplearningmedical}) or yield uncertainty as simplistic by-products, such as the variational inference strategy in \citet{dalca2019unsupervised, sedghi2019probabilistic}. Since these methods often fail to capture the spatial distribution of deformation fields, more recent work has proposed aleatoric formulations tailored to registration \citep{zhang2024heteroscedastic}, but their scope remains limited to voxel-level noise modeling. Recent studies \citep{gopinath2024registration, hu2024hierarchical} have modeled uncertainty hierarchically, tracing its propagation across network outputs, transformation parameters, and downstream tasks. While promising, the existing approaches require architectural modifications and retraining the network, thus not suitable for pretrained models. Furthermore, they rarely provide theoretical clarity on what the predicted variance represents, leaving many uncertainty maps as task-specific heuristics with limited interpretability. Finally, another important gap in existing methods is the limited investigation of how estimated uncertainties correlate with actual registration errors \citep{luo2019applicability, luo2020registration}. This may be partly due to the scarcity of datasets with manual annotations of landmark pairs that could serve as a gold standard. 

In contrast with these methods, we introduce an inference-time, model-agnostic uncertainty estimation for pretrained registration models. This design not only facilitates broad interpretability but also ensures applicability to pretrained registration foundation models such as uniGradICON and SynthMorph, without the need for retraining.

\section{Method}\label{sec:method}
\subsection{Background}\label{sec:background}

Let $f_\theta(I^A, I^B)$ denote a registration network that estimates the transformation $\hat{\phi}^{AB}:\Omega^B\to\Omega^A$ from the domain of the target image $I^B:\Omega^B\,{\to}\,\mathds{R}$ to the source image $I^A:\Omega^A\,{\to}\,\mathds{R}$. Such registration networks are generally trained unsupervisedly over a dataset with the following loss
\begin{equation}
    {L}_{sim}(I^A\circ\hat\phi^{AB}, I^B) + \lambda{L}_{reg}(\hat\phi^{AB})\, , \hat\phi^{AB}=f_\theta(I^A, I^B),
\end{equation} 
where ${L}_{sim}(\cdot, \cdot)$ is a similarity measure between two images (e.g., mean squared error, or normalized cross-correlation), and ${L}_{reg}(\cdot)$ is a regularization term (e.g., smoothness of the transformation).

\textbf{Transformation equivariance for image registration models.} The anatomical correspondence between $I^A$ and $I^B$ is an intrinsic relationship that is independent of the coordinate system in which $I^A$ is sampled (\cref{fig:transformation_equivariance}). Consequently, for any admissible, topology-preserving spatial perturbation $\tau$ (e.g., rigid or diffeomorphic transformations), if $I^A$ undergoes the perturbation $\tau$, resulting in $I^{A'} := I^A \circ \tau$, the mapping between the images should remain consistent after accounting for $\tau$. Formally, a registration model $f(\cdot,\cdot)$ is said to be transformation equivariant if, for any such $\tau$, it satisfies
\begin{equation}\label{eq:transformation_equivariant}
    f(I^A, I^B) = \tau \circ f(I^A \circ \tau, I^B).
\end{equation}

\subsection{The Origin of Registration Uncertainty}\label{sec:intuition}
In practice, transformation equivariance does not occur for several reasons, each reflecting uncertainty in correspondence estimation. First, ambiguity may be introduced at the image acquisition level. Artifacts, noise, or limited resolution can obscure anatomical cues, giving rise to one-to-many correspondences in which multiple plausible mappings are consistent with the observed images. In such cases, no registration method can be expected to preserve equivariance, as different perturbations may lead the model to select different valid correspondences.

Second, even when the anatomical information is preserved by the images, a registration model may fail to extract the correct correspondence due to limited model capacity, suboptimal training, or architectural bias. In this scenario, equivariance violations reflect instability of the model itself rather than ambiguity in the data. 

In both cases, the underlying source of uncertainty lies in the correspondence inferred from the images. We therefore define registration uncertainty as uncertainty in anatomical correspondence, arising from multiple factors. Rather than attempting to disentangle these causes, we propose quantifying uncertainty by measuring the variability of correspondence estimates under equivalent formulations of the same registration task. As discussed above, this variability is naturally captured by the variance of composed transformations under spatial perturbations.

\subsection{Perturbation-Based Uncertainty Estimation}
\label{sec:uncertainty_framework}

We estimate uncertainty by applying random perturbations to the source image and measuring the variance of the composed predictions. Formally, we estimate the uncertainty via 
\vspace{-.05in}
\begin{equation}
\operatorname{Var}_{\tau \sim \mathcal{T}}\!\left[\tau \circ f_\theta(I^A \circ \tau, I^B)\right],
\end{equation}
with $\mathcal{T}$ being a distribution $\tau$ samples from. The perturbation family can be chosen flexibly (e.g., translations, affine maps, elastic deformations), as is the registration network. Our framework is agnostic to these choices. The algorithm is described in \cref{alg:spv}. The per-voxel uncertainty map is computed from the sample variance of the composed outputs. We report the scalar uncertainty $u(y)=\sqrt{\mathrm{tr}\,S(y)}$ (root trace), unless otherwise stated. Implementation details and statistics are provided in \cref{appendix:sec:uncertainty_implement_details}.

\begin{algorithm}
\DontPrintSemicolon
\caption{Perturbation-based Uncertainty Estimation}\label{alg:spv}
\KwIn{Source $I^A$, target $I^B$, network $f_\theta$, samples $N$, perturbation distribution $\mathcal{T}$}
\KwOut{Mean field $\mu(y)$, uncertainty map $u(y)$}
\For{$n=1$ \KwTo $N$}{
    sample $\tau_n \sim \mathcal{T}$\;
    $I^{A'}_n \leftarrow \text{SpatialTransform}(I^A,\tau_n)$\;
    $\hat{\phi}^{A'B}_n \leftarrow f_\theta(I^{A'}_n, I^B)$\;
    $g_n \leftarrow \text{Compose}(\tau_n, \hat{\phi}^{A'B}_n)$\;
}
$\mu(y) \leftarrow \tfrac{1}{N}\sum_n g_n(y)$\;
$S(y) \leftarrow \tfrac{1}{N}\sum_n \|g_n(y)-\mu(y)\|^2$\;
$u(y) \leftarrow \sqrt{\mathrm{tr}\,S(y)}$\;
\Return $\mu(y),\,u(y)$
\end{algorithm}

\section{Theoretical Analysis and Connection with Registration Error}\label{sec:theory_analysis}

 To understand whether the proposed uncertainty estimate can reflect the actual registration error, we decompose the variance with respect to the registration error, providing a theoretical justification for their relationship.

\subsection{Input Perturbation and Composition for Registration Networks \texorpdfstring{$f_\theta(\cdot,\cdot)$}{}}\label{sec:composition_definision}

For a given target point $y \in \Omega^B$, the network predicts its corresponding location in the source space, \emph{approximating} the true transformation $\phi^{AB}$,
\begin{equation}
    \hat{\phi}^{AB}(y):=f_\theta(I^A, I^B)(y) = \phi^{AB}(y) + \epsilon(y),
\end{equation}
where $\epsilon(y)$ represents the error of the network output. Note that we use $\hat{\phi}$ to represent the prediction of the registration network and $\phi$ to represent the true transformation.

Now consider a transformation $\tau \in \text{Diff}(\Omega)$ and $\tau:\Omega^{A^\prime}\to\Omega^A$. Applying the registration network to the perturbed image pair $ (I^{A'}, I^B) $, we obtain
\begin{equation}\label{equ:purturbed_registration}
\hat{\phi}^{A^{\prime}B}(y):=f_\theta(I^A \circ \tau, I^B)(y) = \phi^{A^{\prime}B}(y) + \epsilon_\tau(y).
\end{equation}

As we discussed in \cref{sec:background}, the true mapping between $I^A$ and $I^B$ should remain consistent. Thus, we have 
\begin{equation}\label{equ:registration_equivariance}
    \phi^{A^{\prime}B}(y) = \tau^{-1}\circ\phi^{AB}(y),
\end{equation}
with $\tau$ being diffeomorphic.
Substituting $\phi^{A^{\prime}B}(y)$ in \cref{equ:purturbed_registration} with \cref{equ:registration_equivariance}, we have
\begin{equation}
\hat{\phi}^{A^{\prime}B}(y):=f_\theta(I^A \circ \tau, I^B)(y) = (\tau^{-1}\circ\phi^{AB})(y) + \epsilon_\tau(y).
\end{equation}

By composing the perturbed transformation $\tau$ and the output $\hat{\phi}^{A^{\prime}B}(y)$ of the registration network given the perturbed image pair, we have
\begin{equation}\label{equ:composed_transformation}
    g_\tau(y) := \tau \circ \hat{\phi}^{A^{\prime}B}(y) = \tau((\tau^{-1}\circ\phi^{AB})(y) + \epsilon_\tau(y)).
\end{equation}
In the following section, we analyze the relationship between the mean and variance of $g_\tau(y)$ and the error of the prediction $\epsilon(y)$.

\subsection{Mean and Variance of \texorpdfstring{$g_\tau(y)$}{}}\label{sec:theory_proof}

We now analyze the expected behavior of the composed outputs $g_\tau(y)$ across random perturbations $\tau \sim \mathcal{T}$, under the assumption that the registration error $\epsilon$ follows a distribution characterized by mean and covariance.

\textbf{Error model.} Given a registration output under a perturbed source input,
\begin{equation}\label{equ:error_model}
    \hat{\phi}^{A'B}(y) = (\tau^{-1} \circ \phi^{AB})(y) + \epsilon_\tau(y),
\end{equation}
we assume the \footnote{See \cref{appendix:sec:detailed_proof} for rationale of the additive error model.}{residual error} $\epsilon_{\tau}(y)\in\R^{d}$ follows a distribution $\mathcal{P}$ characterized by mean $\mu_{\!\epsilon}(\tau;y)$ and covariance $\Sigma_{\!\epsilon}(\tau;y)$\footnote{We use $\mu_{\epsilon}(\tau;y)$ and $\Sigma_{\epsilon}(\tau;y)$ to denote the mean and covariance 
of the residual at voxel $y$ under perturbation $\tau$, noting they are equivalent to $\mu_{\epsilon}(y;\tau)$ 
and $\Sigma_{\epsilon}(y;\tau)$.},
\begin{equation}
\epsilon_{\tau}(y)\;\sim\;
\mathcal{P},
\qquad
\mu_{\!\epsilon}(\tau;y)\in\R^{d},\;
\Sigma_{\!\epsilon}(\tau;y)\in\R^{d\times d}.
\end{equation}

We analyze the composed prediction $g_\tau(y)$ under three perturbation classes:  
\emph{(i) general diffeomorphisms},  
\emph{(ii) affine maps}, and  
\emph{(iii) translations}. Detailed proofs are provided in \cref{appendix:sec:detailed_proof}.

\begin{lemma}[Mean and covariance under an arbitrary diffeomorphic perturbation]
\label{lemma:uncertainty_proxy}
Let the perturbed output be
$g_{\tau}(y)
\;=\;\tau\!\left(\,(\tau^{-1}\!\circ \phi^{AB})(y)+\epsilon_\tau(y)\,\right)$, with
$\epsilon_\tau(y)\sim\mathcal N\!\bigl(\mu_{\!\epsilon}(\tau;y),
                                        \Sigma_{\!\epsilon}(\tau;y)\bigr)$,
and denote by
$J_{\tau}(y):=D\tau_{u}\bigl|_{u=\phi^{AB}(y)}\in\R^{d\times d}$
the Jacobian of the perturbation at voxel $y$.
Without assuming any independence between $\tau$ and
$\epsilon_\tau$ we have
\begin{align}
\label{eq:mean_general}
\mathbb{E}_{\tau}\bigl[g_{\tau}(y)\bigr]
    &= \phi^{AB}(y)
       \;+\;
       \mathbb{E}_{\tau}\!\bigl[
           J_{\tau}(y)\,\mu_{\!\epsilon}(\tau;y)
         \bigr],
\\[4pt]
\label{eq:cov_general}
\operatorname{Cov}_{\tau}\!\bigl[g_{\tau}(y)\bigr]
    &= \mathbb{E}_{\tau}\!\bigl[
         J_{\tau}(y)\,
         \Sigma_{\!\epsilon}(\tau;y)\,
         J_{\tau}^{\!\top}(y)
       \bigr] \notag\\
    &\quad\quad\quad\quad\quad +\;
       \operatorname{Cov}_{\tau}\!\bigl[
         J_{\tau}(y)\,\mu_{\!\epsilon}(\tau;y)
       \bigr].
\end{align}
\end{lemma}
\textbf{Interpretation.} The decomposition in \cref{eq:cov_general} reveals that the total measured variance is composed of two distinct geometric components, and we refer to these components as \textit{intrinsic spread} and \textit{bias jitter}.
\begin{itemize}[leftmargin=1.5em,itemsep=4pt]
\item \textbf{Intrinsic Spread.}  
The first term (\textcolor{blue}{\textbf{blue arrow}}; \cref{fig:teaser}) corresponds to the expected covariance of the residual error, $\Sigma_{\epsilon}$, as it is linearly propagated through the local Jacobian $J_{\tau}$ of the perturbation. This component captures the average local variability of the model’s prediction errors across different perturbations, reflecting the stability of the predicted correspondence.

\item \textbf{Bias Jitter.}  
The second term (\textcolor{mypurple}{\textbf{purple arrow}}; \cref{fig:teaser}) captures the covariance of the residual mean $\mu_{\epsilon}$ induced by different perturbations. This component measures the consistency of the model’s systematic error, quantifying how the center of the error distribution shifts when the registration problem is reformulated under different spatial perturbations.
\end{itemize}

\textbf{How does the error model affect the variance?} We model the registration error as a distribution without loss of generality in our analysis. For a deterministic registration network that outputs a single transformation, the error model falls back to a constant, and thus the residual variance vanishes ($\Sigma_{\!\epsilon}=0$), so the perturbation covariance reduces to 
\begin{equation}
    \Cov_\tau\bigl[g_{\tau}(y)\bigr]=\;\Var_\tau\!\bigl[J_\tau(y)\,\mu_{\!\epsilon}(\tau;y)\bigr]\!.
\end{equation}

Thus, the same decomposition still applies, meaning the uncertainty map reflects perturbation-sensitive \textit{bias jitter} even when intrinsic variance is absent.

\begin{lemma}[Mean and variance under affine perturbation]
\label{lemma:affine_error}
Let $ \tau(z) = A z + b $ be a random affine transformation with $ A \in \mathbb{R}^{d \times d} $, $ b \in \mathbb{R}^d $. Since the translation part cancels, $g_{\tau}(y)=\phi^{AB}(y)+A\epsilon_\tau(y)$. Then:
\begin{align}
\label{eq:affine_mean}
\mathbb{E}_{A,b}\!\bigl[g_{A,b}(y)\bigr]
  &= \phi^{AB}(y)\;+\;\mathbb{E}_{A}\!\bigl[A\,\mu_{\!\epsilon}(A;y)\bigr],\\[4pt]
\label{eq:affine_var}
\operatorname{Var}_{A,b}\!\bigl[g_{A,b}(y)\bigr]
  &= \mathbb{E}_{A}\!\bigl[A\,\Sigma_{\!\epsilon}(A;y)\,A^{\!\top}\bigr] \notag\\
    &\quad\quad\quad\quad\quad +\;\operatorname{Var}_{A}\!\bigl[A\,\mu_{\!\epsilon}(A;y)\bigr].
\end{align}
\end{lemma}

\begin{corollary}[Translation perturbations]
\label{lemma:translation}
Translations are the special case of \cref{lemma:affine_error} with $A=I$ and $b=t\sim\mathcal D_t$.  
Then $g_t(y)=\phi^{AB}(y)+\epsilon_t(y)$, and
\begin{align}
\mathbb{E}_{t}[g_t(y)]
 &= \phi^{AB}(y)+\mathbb{E}_{t}\!\bigl[\mu_{\!\epsilon}(t;y)\bigr],\\[4pt]
\operatorname{Var}_{t}[g_t(y)]
 &= \mathbb{E}_{t}\!\bigl[\Sigma_{\!\epsilon}(t;y)\bigr]
    +\operatorname{Var}_{t}\!\bigl[\mu_{\!\epsilon}(t;y)\bigr].
\end{align}
\end{corollary}
In \cref{eq:affine_var}, the first term is the \emph{intrinsic spread} propagated by $A$, and the second is the \emph{bias jitter}, the covariance of residual means under different affines. When $A=I$, the affine perturbation reduces to a pure translation, with the Jacobian equal to the identity.

\section{Experiments}
\label{sec:experiments}

We conduct experiments to compare the proposed uncertainty estimation with uncertainty maps estimated via MC-dropout (\cref{sec:compare_with_mcdropout}) and quantitatively evaluate the  correlation between the proposed uncertainty maps with the registration error maps (\cref{sec:dataset_level_evaluation}). The later experiment is conducted across varying ground-truth transformations (translation, affine, deformation), perturbation types (translation, scale, shear, deformation), anatomical structures (brain, abdomen, cardiac), and registration networks with different backbones. Lastly, we qualitatively analyze two case studies (\cref{sec:exp_case_study}) where uncertainty maps highlight anatomically inconsistent or high-risk regions to justify our observations. Please find more details of metrics, datasets, and backbone networks in \cref{appendix:sec:experiment_settings}.

\textbf{Datasets.} We use a brain MRI dataset curated from 11 public sources (see \cref{appendix:sec:experiment_settings} for details), the IXI brain MRI dataset~\citep{chen2022transmorph,IXI_Dataset}, the ACDC cardiac MRI dataset \citep{bernard2018deep}, and the Learn2Reg abdomen CT dataset \citep{xu2016evaluation} in the quantitative experiments. In addition, we conduct case studies on Brats-Reg \citep{baheti2021brain} and Learn2Reg ThoraxCBCT \citep{hugo2016data, hugo2017longitudinal} datasets. 

\textbf{Backbones.} We conduct experiments on two deterministic foundation models chosen for their strong generalization and one task-specific probabilistic model. The first is uniGradICON \citep{tian2024unigradicon}, a foundation registration model spanning multiple anatomies. The second is the nonrigid registration network from SynthMorph \citep{hoffmann2021synthmorph}, a contrast- and resolution-agnostic registration model for the brain. The third is the probabilistic (MC-dropout) registration model from TransMorph~\citep{chen2022transmorph}. We obtain the above pretrained models from their official GitHub repositories.

\subsection{Comparison with Existing Uncertainty Estimation}
\label{sec:compare_with_mcdropout}

\begin{figure}
    \centering
    \includegraphics[width=\linewidth]{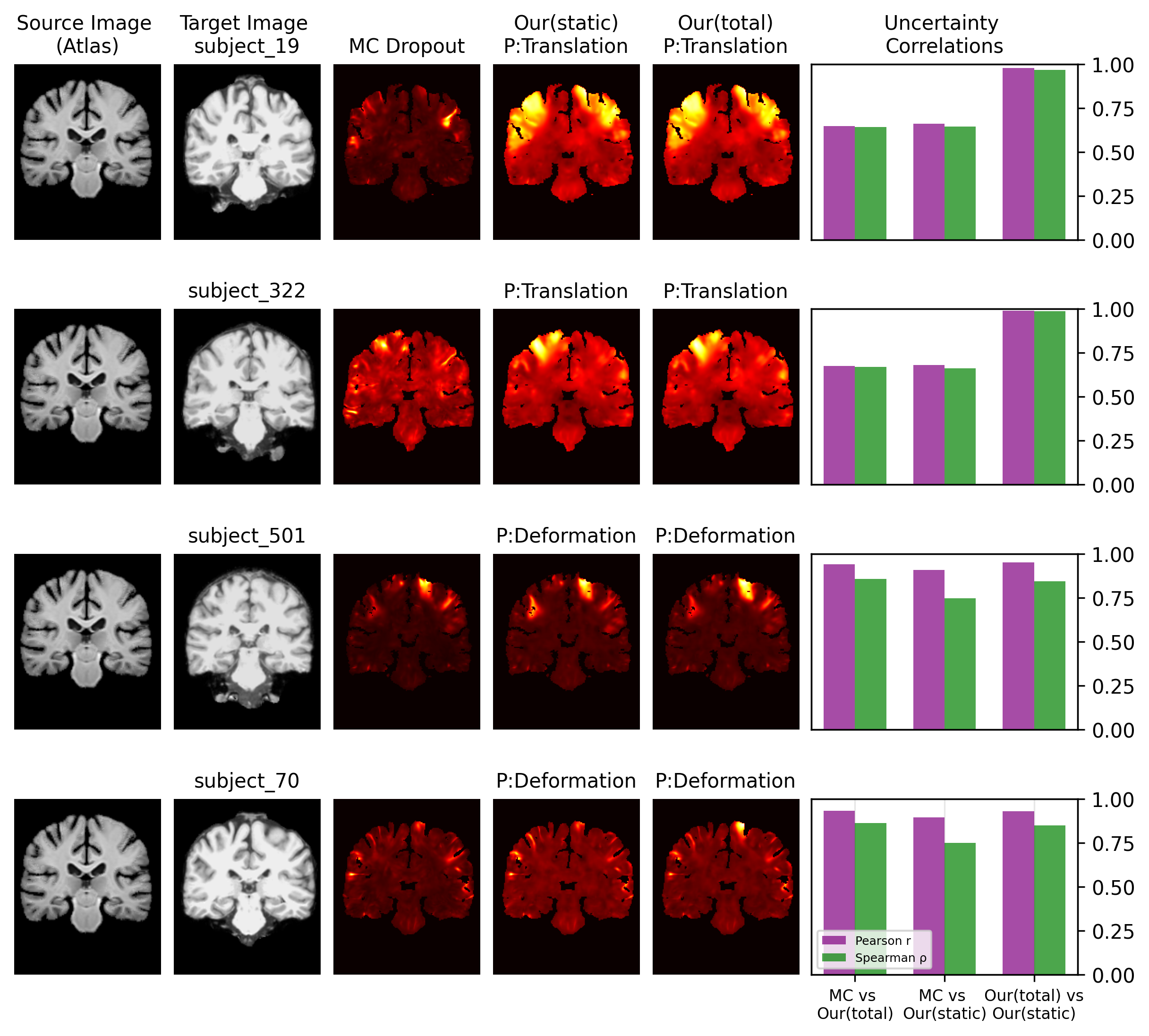}
    \caption{Comparison between our method and the MC-dropout on single cases. The correlation is computed between the 3D uncertainty maps.}
    \label{fig:comparison_with_mc}
    \vspace{-.2in}
\end{figure}

In this experiment, we compare the proposed inference-time uncertainty maps with uncertainty estimates obtained via MC-dropout, a commonly used Bayesian approximation that requires architectural modification and retraining of a deployed model. In contrast, our method operates entirely at inference time and can be applied directly to pretrained registration networks. 

\begin{figure}
    \centering
    \includegraphics[width=\linewidth]{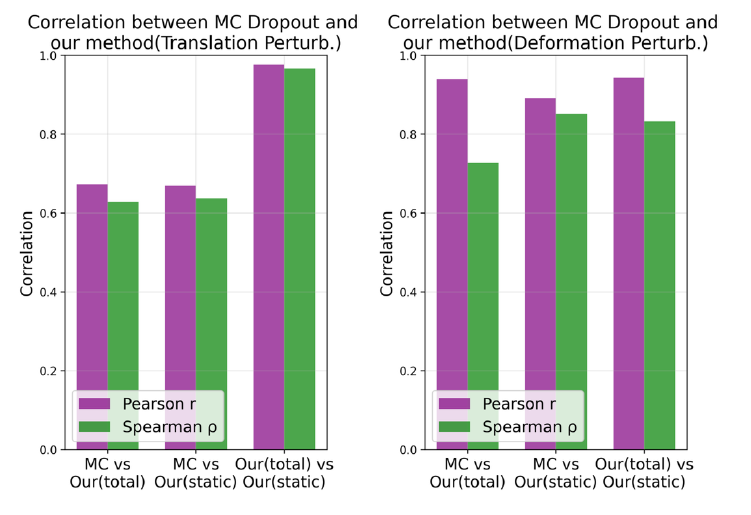}
    \caption{Quantitative comparison between our method and the MC-dropout with the IXI dataset.}
    \label{fig:comparison_with_mc_dataset_level}
    \vspace{-.1in}
\end{figure}

We perform this comparison using the publicly available TransMorph-Bayes model, trained for atlas-to-MRI registration on the IXI dataset. Experiments are conducted on the IXI test set. As no ground-truth uncertainty is available, we assess agreement by computing spatial correlations among three uncertainty maps: (1) MC-dropout uncertainty maps (MC Dropout); (2) proposed uncertainty maps with dropout turned off (Our(static)); and (3) proposed uncertainty maps with dropout on (Our(total)). Figure~\ref{fig:comparison_with_mc} visualizes these uncertainty maps for four representative test cases. Across all examples, the proposed uncertainty maps exhibit strong spatial correlation with MC-dropout uncertainty, indicating that our inference-time estimator captures uncertainty patterns consistent with those produced by retraining-based Bayesian methods. Moreover, we compute the mean of the correlation between the three uncertainty maps across the dataset and show the results in \cref{fig:comparison_with_mc_dataset_level}. The result demonstrates that the strong correlation we observed for single cases is general across the dataset.

\subsection{Quantitative Analysis of the Correlation between the Uncertainty Maps and Registration Error Maps}\label{sec:dataset_level_evaluation}

\begin{figure*}
    \centering
    \includegraphics[width=\linewidth]{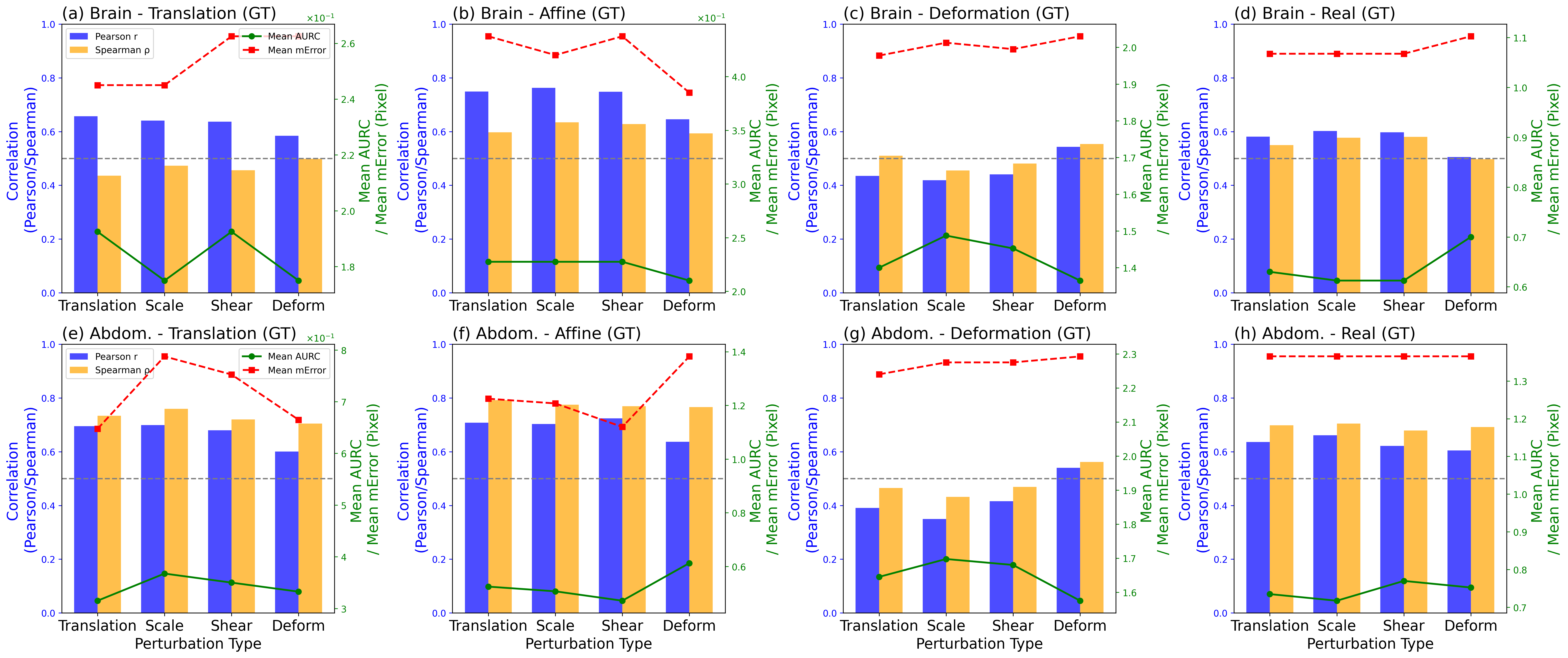}
    \caption{Correlation between the estimated uncertainty and the true registration error for a combination of varying ground truth transformation types (GT), perturbation transformation types, and datasets with uniGradICON.}
    \label{fig:correlation_unigradicon}
    \vspace{-.05in}
\end{figure*}

\begin{figure}
      \includegraphics[width=\linewidth,trim={30.5cm 0 0 0},clip]{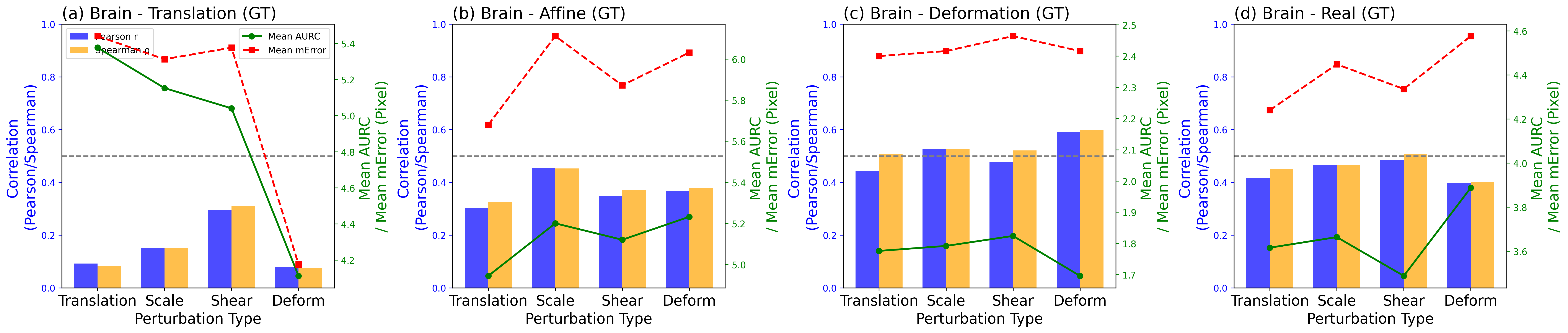}
      \caption{Correlation between the estimated uncertainty and the true registration error for a combination of varying ground truth transformation types (GT) and perturbation transformation types with SynthMorph.}
      \label{fig:correlation_synthmorph}
      \vspace{-.2in}
\end{figure}

To evaluate the correlation between the proposed uncertainty maps and registration maps, we first simulate paired images using three categories of deformations: (i) translations, (ii) affine transformations combining translation, shear, and scale, and (iii) elastic B-spline deformations. While these simulated transformations provide controlled conditions, they do not fully capture the challenges of real registration. To approximate real scenarios while preserving access to ground truth, we also use ANTs \citep{avants2008symmetric} to estimate affine and nonrigid transformations between real image pairs; the estimated transformations are then applied to warp the source image to form new target images. This ensures that the ground-truth mappings are derived from real registrations while remaining accessible for evaluation. Both settings follow the same experimental design, applying perturbations of different types (translation, scale, shear, and deformation) to assess uniGradICON and SynthMorph. Details of the experiment settings are provided in \cref{appendix:sec:dataset_level_evaluation}, and the results are shown in \cref{fig:correlation_unigradicon} and \cref{fig:correlation_synthmorph}. 

For \textbf{uniGradICON}, the estimated uncertainty maps correlates positively with the registration error maps across all settings (\cref{fig:correlation_unigradicon} (a)-(h)). Correlations are strong when the ground truth is linear (translation or affine; \cref{fig:correlation_unigradicon} (a,b,e,f)) and moderate when the ground truth is nonlinear (deformation or real; \cref{fig:correlation_unigradicon} (c,d,g,h)). This drop in correlation likely arises from two interacting factors: (i) nonlinear ground truths induce more localized errors, and (ii) linear perturbations (translation, scale, shear) are less effective at capturing such local discrepancies. See visualizations in \cref{appendix:sec:exp_single_case}. Despite these differences, the AURC (\textcolor{Green}{green line}) consistently improves over the mean error baseline (\textcolor{red}{red line}), confirming that the uncertainty map provides a more informative ranking of errors than chance. 

For \textbf{SynthMorph}, evaluated on brain MRI, strong correlations appear under nonrigid ground truths and real transformations (\cref{fig:correlation_synthmorph} (c, d)) with consistent AURC improvements over the mean mError across settings. This observation is consistent with the observations we had for \textbf{uniGradICON}. In sum, these results demonstrate that perturbation-based uncertainty provides a reliable proxy for registration error. Please refer to \cref{appendix:sec:dataset_level_evaluation} for more results and a detailed discussion of why and when the uncertainty map correlates closely with the error map.

\subsection{Case Study: Uncertainty Maps with Anatomical Inconsistencies}
\label{sec:exp_case_study}

\begin{figure*}
    \centering
    \begin{subfigure}[ht]{\textwidth}
    \includegraphics[width=\linewidth]{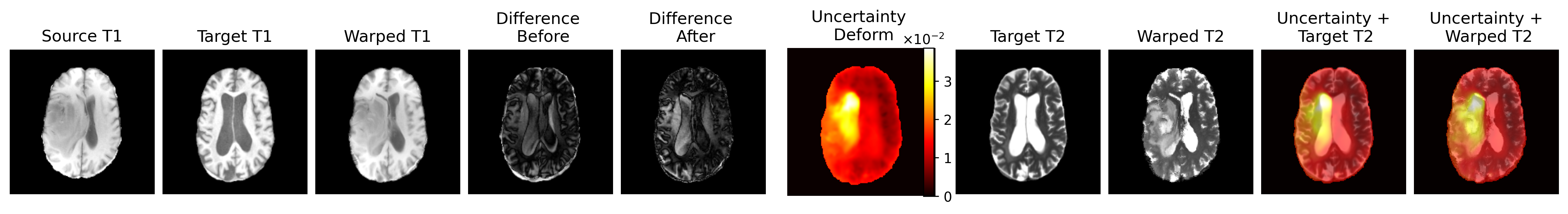}
    \caption{Registration between pre-operative and follow-up MRIs. The highlighted region aligns with the tumor area.}
    \label{fig:exp_case_study_one_bratsreg}
\end{subfigure}

\begin{subfigure}[ht]{\textwidth}
    \includegraphics[width=\linewidth]{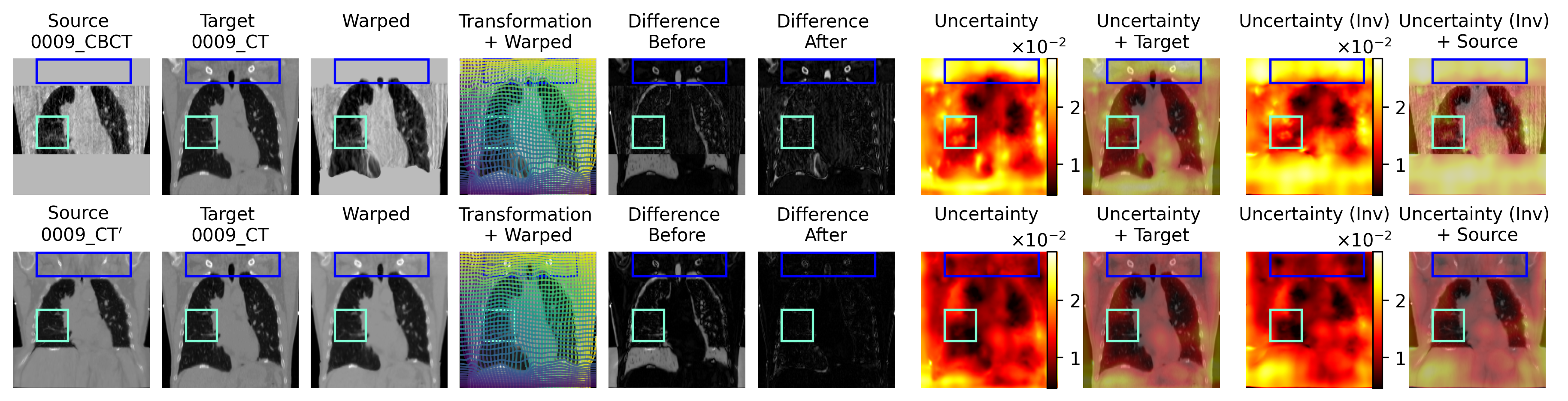}
    \caption{Registration between CBCT and CT, which have different FOVs. Compared to the synthetic CT/CT control pair with matched anatomy, the CBCT/CT setting shows elevated uncertainty in regions outside the CBCT field of view (\textcolor{Blue}{blue box}) and in noisy regions characteristic of CBCT (\textcolor{Aquamarine}{light blue box}). Uncertainty (Inv) denotes the uncertainty map warped via the inverse of $\phi^{AB}$.}
    \label{fig:exp_case_study_two_different_fov}
\end{subfigure}
    \caption{Case studies of uncertainty maps used when there exist anatomical inconsistencies.}
    \label{fig:placeholder}
\end{figure*}

Beyond controlled perturbations, a key question is whether the proposed uncertainty map can highlight critical cases where anatomical correspondence is absent. We explore two case studies. We present one example for each case study and more examples are provided in \cref{appendix:sec:exp_case_study}. 

\textbf{Tumor resection.} Using one pre-/post-operative pair from BraTS-Reg~\citep{baheti2021brain}, we examine whether the proposed uncertainty reflects structural changes induced by surgery. Registrations and corresponding uncertainty maps are computed from T1-weighted images, while T2-weighted images are used for visualization to better delineate abnormal regions. As shown in \cref{fig:exp_case_study_one_bratsreg}, overlaying the uncertainty map on the target and warped T2 reveals that regions of high uncertainty spatially overlap with the tumor area. This behavior is expected, as the tumor region shows barely discriminative anatomical visual cues, making the correspondence between the source and target T1 images inherently ambiguous in this region. 

\textbf{Different fields of view (FOV).} We next examine whether the proposed uncertainty map can highlight regions with missing correspondence caused by mismatched fields of view across imaging modalities (e.g., CBCT and CT). We use data from the Learn2Reg ThoraxCBCT dataset \citep{hugo2016data, hugo2017longitudinal}. Specifically, we register one CBCT/CT pair and compute the corresponding uncertainty map. As a control case, we generate a synthetic source image $CT^\prime$ by warping the CT using the transformation estimated by uniGradICON during CBCT/CT registration. We then register $CT^\prime$ to the original CT, yielding a setting in which no FOV-related inconsistencies are present. The results of the origianl CBCT/CT registration and the control case are shown in \cref{fig:exp_case_study_two_different_fov}. Compared to the synthetic CT/CT control pair with matched anatomy, the uncertainty map of the CBCT/CT registration exhibits elevated values in regions outside the CBCT FOV (\textcolor{Blue}{blue box}), indicating voxels with missing correspondence. In addition, CBCT images contain stronger acquisition artifacts, which obscure anatomical cues and introduce correspondence ambiguity. Consistent with this, we observe increased uncertainty around vascular structures affected by CBCT artifacts (\textcolor{Aquamarine}{light blue box}).

Both case studies demonstrate that the proposed uncertainty map serves as a practical quality-control tool, as it not only correlates with registration error, but also flags clinically relevant correspondence ambiguity. 

\section{Limitations}
Our study has several limitations. The theoretical analysis relies on first- and second-order moments of the error model, leaving higher-order characteristics such as skewness or kurtosis unexplored. Second, the proposed uncertainty does not disentangle different sources of uncertainty in the manner of Bayesian approaches; further analysis from a Bayesian perspective may provide additional insight. Finally, our study focuses on pretrained registration networks, and we have not evaluated the applicability of the proposed method to conventional optimization-based registration algorithms. 

\section{Conclusion}
We introduced an inference-time uncertainty estimation framework for pretrained image registration models based on transformation equivariance, together with a theoretical analysis connecting the uncertainty maps to registration error. Without requiring architectural modifications or retraining, the proposed approach shows consistent agreement with established Bayesian methods such as MC-dropout. Across four anatomical structures and three pretrained registration models, the proposed uncertainty maps exhibit strong to moderate correlation with registration error maps. In addition, the method produces spatially resolved uncertainty maps that highlight anatomically inconsistent regions in representative case studies. By translating variability under spatial perturbations into risk signals, the framework enables practical uncertainty-aware registration for pretrained models for clinical and large-scale research settings.

\clearpage

\section*{Acknowledgements}
This research was funded by NIH grants 1R01EB031114, 1UM1MH130981,  1RF1MH123195, 1RF1AG080371, 1R01AG070988, and 1R21NS138995. The work expresses the views of the authors, not of NIH.

\section*{Impact Statements}
This work focuses on developing an inference-time uncertainty estimation framework for pretrained medical image registration models. All experiments are conducted on publicly available, de-identified datasets. No new human or animal data were collected, and therefore no additional IRB approval was required.

The proposed methodology is intended as a research contribution toward improving the safety and reliability of registration in clinical and large-scale biomedical research settings. In particular, the uncertainty maps are designed to provide interpretable indicators of registration reliability, thereby reducing the risk of unintended misuse of misaligned images in downstream tasks. The method is not intended for direct clinical deployment without further validation in prospective studies.

\bibliography{icml2026}

@article{iglesias2013improved,
  title={Improved inference in Bayesian segmentation using Monte Carlo sampling: Application to hippocampal subfield volumetry},
  author={Iglesias, Juan Eugenio and Sabuncu, Mert Rory and Van Leemput, Koen and Alzheimer’s Disease Neuroimaging Initiative and others},
  journal={MedIA},
  year={2013},
}

@article{avants2008symmetric,
  title={Symmetric diffeomorphic image registration with cross-correlation: evaluating automated labeling of elderly and neurodegenerative brain},
  author={Avants, Brian B and Epstein, Charles L and Grossman, Murray and Gee, James C},
  journal={MedIA},
  year={2008},
}

@article{klein2009elastix,
  title={Elastix: a toolbox for intensity-based medical image registration},
  author={Klein, Stefan and Staring, Marius and Murphy, Keelin and Viergever, Max A and Pluim, Josien PW},
  journal={TMI},
  year={2009},
}

@article{modat2010fast,
  title={Fast free-form deformation using graphics processing units},
  author={Modat, Marc and Ridgway, Gerard R and Taylor, Zeike A and Lehmann, Manja and Barnes, Josephine and Hawkes, David J and others},
  journal={Computer methods and programs in biomedicine},
  year={2010},
}

@inproceedings{heinrich2012globally,
  title={Globally optimal deformable registration on a minimum spanning tree using dense displacement sampling},
  author={Heinrich, Mattias P and Jenkinson, Mark and Brady, Sir Michael and Schnabel, Julia A},
  booktitle={MICCAI},
  year={2012},
}

@article{yang2017quicksilver,
  title={Quicksilver: Fast predictive image registration--a deep learning approach},
  author={Yang, Xiao and Kwitt, Roland and Styner, Martin and Niethammer, Marc},
  journal={NeuroImage},
  year={2017},
}

@inproceedings{cao2017deformable,
  title={Deformable image registration based on similarity-steered {CNN} regression},
  author={Cao, Xiaohuan and Yang, Jianhua and Zhang, Jun and Nie, Dong and Kim, Minjeong and Wang, Qian and Shen, Dinggang},
  booktitle={MICCAI},
  year={2017},
}

@inproceedings{sokooti2017nonrigid,
  title={Nonrigid image registration using multi-scale {3D} convolutional neural networks},
  author={Sokooti, Hessam and De Vos, Bob and Berendsen, Floris and Lelieveldt, Boudewijn PF and I{\v{s}}gum, Ivana and Staring, Marius},
  booktitle={MICCAI},
  year={2017},
}

@inproceedings{de2017end,
  title={End-to-end unsupervised deformable image registration with a convolutional neural network},
  author={De Vos, Bob D and Berendsen, Floris F and Viergever, Max A and Staring, Marius and I{\v{s}}gum, Ivana},
  booktitle={DLMIA/MICCAI},
  year={2017},
}

@article{balakrishnan2019voxelmorph,
  title={{VoxelMorph}: a learning framework for deformable medical image registration},
  author={Balakrishnan, Guha and Zhao, Amy and Sabuncu, Mert R and Guttag, John and Dalca, Adrian V},
  journal={TMI},
  year={2019},
}

@inproceedings{shen2019networks,
  title={Networks for joint affine and non-parametric image registration},
  author={Shen, Zhengyang and Han, Xu and Xu, Zhenlin and Niethammer, Marc},
  booktitle={CVPR},
  year={2019}
}

@inproceedings{niethammer2019metric,
  title={Metric learning for image registration},
  author={Niethammer, Marc and Kwitt, Roland and Vialard, Francois-Xavier},
  booktitle={CVPR},
  year={2019}
}

@inproceedings{mok2020fast,
  title={Fast symmetric diffeomorphic image registration with convolutional neural networks},
  author={Mok, Tony CW and Chung, Albert},
  booktitle={CVPR},
  year={2020}
}

@article{chen2022transmorph,
  title={Transmorph: Transformer for unsupervised medical image registration},
  author={Chen, Junyu and Frey, Eric C and He, Yufan and Segars, William P and Li, Ye and Du, Yong},
  journal={MedIA},
  year={2022},
}

@inproceedings{hering2019mlvirnet,
  title={mlvirnet: Multilevel variational image registration network},
  author={Hering, Alessa and van Ginneken, Bram and Heldmann, Stefan},
  booktitle={MICCAI},
  year={2019},
}

@article{de2019deep,
  title={A deep learning framework for unsupervised affine and deformable image registration},
  author={De Vos, Bob D and Berendsen, Floris F and Viergever, Max A and Sokooti, Hessam and Staring, Marius and I{\v{s}}gum, Ivana},
  journal={MedIA},
  year={2019},
}

@inproceedings{mok2020large,
  title={Large deformation diffeomorphic image registration with laplacian pyramid networks},
  author={Mok, Tony CW and Chung, Albert CS},
  booktitle={MICCAI},
  year={2020},
}

@article{tian2023same++,
  title={SAME++: A Self-supervised Anatomical eMbeddings Enhanced medical image registration framework using stable sampling and regularized transformation},
  author={Tian, Lin and Li, Zi and Liu, Fengze and Bai, Xiaoyu and Ge, Jia and Lu, Le and Niethammer, Marc and Ye, Xianghua and Yan, Ke and Jin, Daikai},
  journal={arXiv:2311.14986},
  year={2023}
}

@article{mok2024modality,
  title={Modality-Agnostic Structural Image Representation Learning for Deformable Multi-Modality Medical Image Registration},
  author={Mok, Tony CW and Li, Zi and Bai, Yunhao and Zhang, Jianpeng and Liu, Wei and Zhou, Yan-Jie and others},
  journal={arXiv:2402.18933},
  year={2024}
}

@article{gal2015bayesian,
  title={Bayesian convolutional neural networks with Bernoulli approximate variational inference},
  author={Gal, Yarin and Ghahramani, Zoubin},
  journal={arXiv preprint arXiv:1506.02158},
  year={2015}
}

@inproceedings{lakshminarayanan2017simple,
  title={Simple and scalable predictive uncertainty estimation using deep ensembles},
  author={Lakshminarayanan, Balaji and Pritzel, Alexander and Blundell, Charles},
  booktitle={NeurIPS},
  year={2017}
}

@article{dalca2019unsupervised,
  title={Unsupervised learning of probabilistic diffeomorphic registration for images and surfaces},
  author={Dalca, Adrian V and Balakrishnan, Guha and Guttag, John and Sabuncu, Mert R},
  journal={MedIA},
  year={2019},
}

@inproceedings{ayhan2018test,
  title={Test-time data augmentation for estimation of heteroscedastic aleatoric uncertainty in deep neural networks},
  author={Ayhan, Murat Seckin and Berens, Philipp},
  booktitle={MIDL},
  year={2018}
}

@article{wang2019aleatoric,
  title={Aleatoric uncertainty estimation with test-time augmentation for medical image segmentation with convolutional neural networks},
  author={Wang, Guotai and Li, Wenqi and Aertsen, Michael and Deprest, Jan and Ourselin, S{\'e}bastien and Vercauteren, Tom},
  journal={Neurocomputing},
  year={2019},
}

@article{di2014autism,
  title={The autism brain imaging data exchange: towards a large-scale evaluation of the intrinsic brain architecture in autism},
  author={Di Martino, Adriana and Yan, Chao-Gan and Li, Qingyang and Denio, Erin and Castellanos, Francisco X and Alaerts, Kaat and Anderson, Jeffrey S and Assaf, Michal and Bookheimer, Susan Y and Dapretto, Mirella and others},
  journal={Molecular psychiatry},
  year={2014},
}

@article{brown2012adhd,
  title={ADHD-200 Global Competition: diagnosing ADHD using personal characteristic data can outperform resting state fMRI measurements},
  author={Brown, Matthew RG and Sidhu, Gagan S and Greiner, Russell and Asgarian, Nasimeh and Bastani, Meysam and Silverstone, Peter H and Greenshaw, Andrew J and Dursun, Serdar M},
  journal={Frontiers in systems neuroscience},
  year={2012},
}

@article{jack2008alzheimer,
  title={The Alzheimer's disease neuroimaging initiative (ADNI): MRI methods},
  author={Jack Jr, Clifford R and Bernstein, Matt A and Fox, Nick C and Thompson, Paul and Alexander, Gene and Harvey, Danielle and Borowski, Bret and Britson, Paula J and L. Whitwell, Jennifer and Ward, Chadwick and others},
  journal={Journal of Magnetic Resonance Imaging: An Official Journal of the International Society for Magnetic Resonance in Medicine},
  year={2008},
}

@article{weiner2017alzheimer,
  title={The Alzheimer's Disease Neuroimaging Initiative 3: Continued innovation for clinical trial improvement},
  author={Weiner, Michael W and Veitch, Dallas P and Aisen, Paul S and Beckett, Laurel A and Cairns, Nigel J and Green, Robert C and Harvey, Danielle and Jack Jr, Clifford R and Jagust, William and Morris, John C and others},
  journal={Alzheimer's \& Dementia},
  year={2017},
}

@article{fowler2021fifteen,
  title={Fifteen years of the Australian Imaging, Biomarkers and Lifestyle (AIBL) study: progress and observations from 2,359 older adults spanning the spectrum from cognitive normality to Alzheimer’s disease},
  author={Fowler, Christopher and Rainey-Smith, Stephanie R and Bird, Sabine and Bomke, Julia and Bourgeat, Pierrick and Brown, Belinda M and Burnham, Samantha C and Bush, Ashley I and Chadunow, Carolyn and Collins, Steven and others},
  journal={Journal of Alzheimer's disease reports},
  year={2021},
}

@article{fischl2002whole,
  title={Whole brain segmentation: automated labeling of neuroanatomical structures in the human brain},
  author={Fischl, Bruce and Salat, David H and Busa, Evelina and Albert, Marilyn and Dieterich, Megan and Haselgrove, Christian and Van Der Kouwe, Andre and Killiany, Ron and Kennedy, David and Klaveness, Shuna and others},
  journal={Neuron},
  year={2002},
}

@article{mayer2013functional,
  title={Functional imaging of the hemodynamic sensory gating response in schizophrenia},
  author={Mayer, Andrew R and Ruhl, David and Merideth, Flannery and Ling, Josef and Hanlon, Faith M and Bustillo, Juan and Canive, Jose},
  journal={Human brain mapping},
  year={2013},
}

@article{vogt2023chinese,
  title={The Chinese Human Connectome Project},
  author={Vogt, Nina},
  journal={Nature Methods},
  year={2023},
}

@article{van2012human,
  title={The Human Connectome Project: a data acquisition perspective},
  author={Van Essen, David C and Ugurbil, Kamil and Auerbach, Edward and Barch, Deanna and Behrens, Timothy EJ and Bucholz, Richard and Chang, Acer and Chen, Liyong and Corbetta, Maurizio and Curtiss, Sandra W and others},
  journal={Neuroimage},
  year={2012},
}

@article{carass2017longitudinal,
  title={Longitudinal multiple sclerosis lesion segmentation: resource and challenge},
  author={Carass, Aaron and Roy, Snehashis and Jog, Amod and Cuzzocreo, Jennifer L and Magrath, Elizabeth and Gherman, Adrian and Button, Julia and Nguyen, James and Prados, Ferran and Sudre, Carole H and others},
  journal={NeuroImage},
  year={2017},
}

@article{gollub2013mcic,
  title={The MCIC collection: a shared repository of multi-modal, multi-site brain image data from a clinical investigation of schizophrenia},
  author={Gollub, Randy L and Shoemaker, Jody M and King, Margaret D and White, Tonya and Ehrlich, Stefan and Sponheim, Scott R and Clark, Vincent P and Turner, Jessica A and Mueller, Bryon A and Magnotta, Vince and others},
  journal={Neuroinformatics},
  year={2013},
}

@article{lamontagne2019oasis,
  title={OASIS-3: longitudinal neuroimaging, clinical, and cognitive dataset for normal aging and Alzheimer disease},
  author={LaMontagne, Pamela J and Benzinger, Tammie LS and Morris, John C and Keefe, Sarah and Hornbeck, Russ and Xiong, Chengjie and Grant, Elizabeth and Hassenstab, Jason and Moulder, Krista and Vlassenko, Andrei G and others},
  journal={medrxiv},
  year={2019},
}

@article{baheti2021brain,
  title={The brain tumor sequence registration (brats-reg) challenge: Establishing correspondence between pre-operative and follow-up mri scans of diffuse glioma patients},
  author={Baheti, Bhakti and Chakrabarty, Satrajit and Akbari, Hamed and Bilello, Michel and Wiestler, Benedikt and Schwarting, Julian and Calabrese, Evan and Rudie, Jeffrey and Abidi, Syed and Mousa, Mina and others},
  journal={arXiv preprint arXiv:2112.06979},
  year={2021}
}

@misc{hugo2016data,
  doi = {10.7937/K9/TCIA.2016.ELN8YGLE},
  url = {https://www.cancerimagingarchive.net/collection/4d-lung/},
  author = {Hugo,  Geoffrey D. and Weiss,  Elisabeth and Sleeman,  William C. and Balik,  Salim and Keall,  Paul J. and Lu,  Jun and Williamson,  Jeffrey F.},
  title = {Data from 4D Lung Imaging of NSCLC Patients},
  year = {2016},
  copyright = {Creative Commons Attribution 3.0 Unported}
}

@article{hugo2017longitudinal,
  title={A longitudinal four-dimensional computed tomography and cone beam computed tomography dataset for image-guided radiation therapy research in lung cancer},
  author={Hugo, Geoffrey D and Weiss, Elisabeth and Sleeman, William C and Balik, Salim and Keall, Paul J and Lu, Jun and Williamson, Jeffrey F},
  journal={Medical physics},
  year={2017},
}

@inproceedings{tian2024unigradicon,
  title={unigradicon: A foundation model for medical image registration},
  author={Tian, Lin and Greer, Hastings and Kwitt, Roland and Vialard, Francois-Xavier and San Jos{\'e} Est{\'e}par, Ra{\'u}l and Bouix, Sylvain and Rushmore, Richard and Niethammer, Marc},
  booktitle={MICCAI},
  year={2024},
}

@article{hoffmann2021synthmorph,
  title={SynthMorph: learning contrast-invariant registration without acquired images},
  author={Hoffmann, Malte and Billot, Benjamin and Greve, Douglas N and Iglesias, Juan Eugenio and Fischl, Bruce and Dalca, Adrian V},
  journal={TMI},
  year={2021},
}

@article{xu2016evaluation,
  title={Evaluation of six registration methods for the human abdomen on clinically acquired CT},
  author={Xu, Zhoubing and Lee, Christopher P and Heinrich, Mattias P and Modat, Marc and Rueckert, Daniel and Ourselin, Sebastien and Abramson, Richard G and Landman, Bennett A},
  journal={TBME},
  year={2016},
}

@article{bernard2018deep,
  title={Deep learning techniques for automatic MRI cardiac multi-structures segmentation and diagnosis: is the problem solved?},
  author={Bernard, Olivier and Lalande, Alain and Zotti, Clement and Cervenansky, Frederick and Yang, Xin and Heng, Pheng-Ann and Cetin, Irem and Lekadir, Karim and Camara, Oscar and Ballester, Miguel Angel Gonzalez and others},
  journal={TMI},
  year={2018},
}

@inproceedings{simpson2013bayesian,
  title={A bayesian approach for spatially adaptive regularisation in non-rigid registration},
  author={Simpson, Ivor JA and Woolrich, Mark W and Cardoso, Manuel Jorge and Cash, David M and Modat, Marc and Schnabel, Julia A and Ourselin, Sebastien},
  booktitle={MICCAI},
  year={2013},
}

@article{kybic2009bootstrap,
  title={Bootstrap resampling for image registration uncertainty estimation without ground truth},
  author={Kybic, Jan},
  journal={TIP},
  year={2009},
}

@article{le2017sparse,
  title={Sparse Bayesian registration of medical images for self-tuning of parameters and spatially adaptive parametrization of displacements},
  author={Le Folgoc, Lo{\"\i}c and Delingette, Herv{\'e} and Criminisi, Antonio and Ayache, Nicholas},
  journal={MedIA},
  year={2017},
}

@article{risholm2013bayesian,
  title={Bayesian characterization of uncertainty in intra-subject non-rigid registration},
  author={Risholm, Petter and Janoos, Firdaus and Norton, Isaiah and Golby, Alex J and Wells III, William M},
  journal={MedIA},
  year={2013},
}

@article{le2016quantifying,
  title={Quantifying registration uncertainty with sparse bayesian modelling},
  author={Le Folgoc, Loic and Delingette, Herve and Criminisi, Antonio and Ayache, Nicholas},
  journal={TMI},
  year={2016},
}

@inproceedings{agn2019fast,
  title={Fast nonparametric mutual-information-based registration and uncertainty estimation},
  author={Agn, Mikael and Van Leemput, Koen},
  booktitle={Uncertainty for Safe Utilization of Machine Learning in Medical Imaging and Clinical Image-Based Procedures: First International Workshop, UNSURE 2019, and 8th International Workshop, CLIP 2019, Held in Conjunction with MICCAI 2019},
  year={2019},
}

@inproceedings{gong2022uncertainty,
  title={Uncertainty learning towards unsupervised deformable medical image registration},
  author={Gong, Xuan and Khaidem, Luckyson and Zhu, Wentao and Zhang, Baochang and Doermann, David},
  booktitle={WACV},
  year={2022}
}

@inproceedings{smolders2022deformable,
  title={Deformable image registration uncertainty quantification using deep learning for dose accumulation in adaptive proton therapy},
  author={Smolders, Andreas and Lomax, Tony and Weber, Damien Charles and Albertini, Francesca},
  booktitle={WBIR},
  year={2022},
}

@article{chen2024registration,
  title={From Registration Uncertainty to Segmentation Uncertainty},
  author={Chen, Junyu and Liu, Yihao and Wei, Shuwen and Bian, Zhangxing and Carass, Aaron and Du, Yong},
  journal={arXiv preprint arXiv:2403.05111},
  year={2024}
}

@misc{chen2024surveydeeplearningmedical,
      title={A survey on deep learning in medical image registration: new technologies, uncertainty, evaluation metrics, and beyond}, 
      author={Junyu Chen and Yihao Liu and Shuwen Wei and Zhangxing Bian and Shalini Subramanian and Aaron Carass and Jerry L. Prince and Yong Du},
      year={2024},
      eprint={2307.15615},
      archivePrefix={arXiv},
      primaryClass={eess.IV},
      url={https://arxiv.org/abs/2307.15615}, 
}

@inproceedings{sedghi2019probabilistic,
  title={Probabilistic image registration via deep multi-class classification: characterizing uncertainty},
  author={Sedghi, Alireza and Kapur, Tina and Luo, Jie and Mousavi, Parvin and Wells, William M},
  booktitle={Uncertainty for Safe Utilization of Machine Learning in Medical Imaging and Clinical Image-Based Procedures: First International Workshop, UNSURE 2019, and 8th International Workshop, CLIP 2019, Held in Conjunction with MICCAI 2019},
  year={2019},
}

@inproceedings{luo2019applicability,
  title={On the applicability of registration uncertainty},
  author={Luo, Jie and Sedghi, Alireza and Popuri, Karteek and Cobzas, Dana and Zhang, Miaomiao and Preiswerk, Frank and Toews, Matthew and Golby, Alexandra and Sugiyama, Masashi and Wells, William M and others},
  booktitle={MICCAI},
  year={2019},
}

@inproceedings{zhang2024heteroscedastic,
  title={Heteroscedastic uncertainty estimation framework for unsupervised registration},
  author={Zhang, Xiaoran and Pak, Daniel H and Ahn, Shawn S and Li, Xiaoxiao and You, Chenyu and Staib, Lawrence H and Sinusas, Albert J and Wong, Alex and Duncan, James S},
  booktitle={MICCAI},
  year={2024},
}

@inproceedings{luo2020registration,
  title={Are registration uncertainty and error monotonically associated?},
  author={Luo, Jie and Frisken, Sarah and Wang, Duo and Golby, Alexandra and Sugiyama, Masashi and Wells III, William},
  booktitle={MICCAI},
  year={2020},
}

@article{leung2010automated,
  title={Automated cross-sectional and longitudinal hippocampal volume measurement in mild cognitive impairment and Alzheimer's disease},
  author={Leung, Kelvin K and Barnes, Josephine and Ridgway, Gerard R and Bartlett, Jonathan W and Clarkson, Matthew J and Macdonald, Kate and Schuff, Norbert and Fox, Nick C and Ourselin, Sebastien and Alzheimer's Disease Neuroimaging Initiative and others},
  journal={Neuroimage},
  year={2010},
}

@article{tustison2019longitudinal,
  title={Longitudinal mapping of cortical thickness measurements: An Alzheimer’s Disease Neuroimaging Initiative-based evaluation study},
  author={Tustison, Nicholas J and Holbrook, Andrew J and Avants, Brian B and Roberts, Jared M and Cook, Philip A and Reagh, Zachariah M and Duda, Jeffrey T and Stone, James R and Gillen, Daniel L and Yassa, Michael A and others},
  journal={Journal of Alzheimer’s Disease},
  year={2019},
  publisher={SAGE Publications Sage UK: London, England}
}

@inproceedings{rokuss2025lesionlocator,
  title={LesionLocator: Zero-Shot Universal Tumor Segmentation and Tracking in 3D Whole-Body Imaging},
  author={Rokuss, Maximilian and Kirchhoff, Yannick and Akbal, Seval and Kovacs, Balint and Roy, Saikat and Ulrich, Constantin and Wald, Tassilo and Rotkopf, Lukas T and Schlemmer, Heinz-Peter and Maier-Hein, Klaus},
  booktitle={CVPR},
  year={2025}
}

@article{kessler2006image,
  title={Image registration and data fusion in radiation therapy},
  author={Kessler, Marc L},
  journal={The British journal of radiology},
  year={2006},
}

@inproceedings{gopinath2024registration,
  title={Registration by regression (rbr): a framework for interpretable and flexible atlas registration},
  author={Gopinath, Karthik and Hu, Xiaoling and Hoffmann, Malte and Puonti, Oula and Iglesias, Juan Eugenio},
  booktitle={WBIR},
  year={2024},
}

@inproceedings{hu2024hierarchical,
  title={Hierarchical uncertainty estimation for learning-based registration in neuroimaging},
  author={Hu, Xiaoling and Gopinath, Karthik and Liu, Peirong and Hoffmann, Malte and Van Leemput, Koen and Puonti, Oula and Iglesias, Juan Eugenio},
  booktitle={ICLR},
  year={2025}
}

@article{dubost2020multi,
  title={Multi-atlas image registration of clinical data with automated quality assessment using ventricle segmentation},
  author={Dubost, Florian and de Bruijne, Marleen and Nardin, Marco and Dalca, Adrian V and Donahue, Kathleen L and Giese, Anne-Katrin and Etherton, Mark R and Wu, Ona and de Groot, Marius and Niessen, Wiro and others},
  journal={MedIA},
  year={2020},
}

@inproceedings{rupprecht2017learning,
  title={Learning in an uncertain world: Representing ambiguity through multiple hypotheses},
  author={Rupprecht, Christian and Laina, Iro and DiPietro, Robert and Baust, Maximilian and Tombari, Federico and Navab, Nassir and Hager, Gregory D},
  booktitle={ICCV},
  year={2017}
}

@inproceedings{moon2020confidence,
  title={Confidence-aware learning for deep neural networks},
  author={Moon, Jooyoung and Kim, Jihyo and Shin, Younghak and Hwang, Sangheum},
  booktitle={ICML},
  year={2020},
}

@inproceedings{li2023confidence,
  title={Confidence estimation using unlabeled data},
  author={Li, Chen and Hu, Xiaoling and Chen, Chao},
  booktitle={ICLR},
  year={2023}
}

@inproceedings{tian2023gradicon,
  title={GradICON: Approximate diffeomorphisms via gradient inverse consistency},
  author={Tian, Lin and Greer, Hastings and Vialard, Fran{\c{c}}ois-Xavier and Kwitt, Roland and Est{\'e}par, Ra{\'u}l San Jos{\'e} and Rushmore, Richard Jarrett and Makris, Nikolaos and Bouix, Sylvain and Niethammer, Marc},
  booktitle={CVPR},
  year={2023}
}

@inproceedings{greer2021icon,
  title={ICON: Learning regular maps through inverse consistency},
  author={Greer, Hastings and Kwitt, Roland and Vialard, Fran{\c{c}}ois-Xavier and Niethammer, Marc},
  booktitle={ICCV},
  year={2021}
}

@inproceedings{tian2024nephi,
  title={Nephi: Neural deformation fields for approximately diffeomorphic medical image registration},
  author={Tian, Lin and Greer, Hastings and Jos{\'e} Est{\'e}par, Ra{\'u}l San and Sengupta, Roni and Niethammer, Marc},
  booktitle={ECCV},
  year={2024},
}

@inproceedings{song2024dino,
  title={Dino-reg: General purpose image encoder for training-free multi-modal deformable medical image registration},
  author={Song, Xinrui and Xu, Xuanang and Yan, Pingkun},
  booktitle={MICCAI},
  year={2024},
}

@misc{IXI_Dataset,
  url = {https://brain-development.org/ixi-dataset/}
}

@article{hering2022learn2reg,
  title={Learn2Reg: comprehensive multi-task medical image registration challenge, dataset and evaluation in the era of deep learning},
  author={Hering, Alessa and Hansen, Lasse and Mok, Tony CW and Chung, Albert CS and Siebert, Hanna and H{\"a}ger, Stephanie and Lange, Annkristin and Kuckertz, Sven and Heldmann, Stefan and Shao, Wei and others},
  journal={IEEE Transactions on Medical Imaging},
  volume={42},
  number={3},
  pages={697--712},
  year={2022},
  publisher={IEEE}
}

@article{chen2025beyond,
  title={Beyond the LUMIR challenge: The pathway to foundational registration models},
  author={Chen, Junyu and Wei, Shuwen and Honkamaa, Joel and Marttinen, Pekka and Zhang, Hang and Liu, Min and Zhou, Yichao and Tan, Zuopeng and Wang, Zhuoyuan and Wang, Yi and others},
  journal={arXiv preprint arXiv:2505.24160},
  year={2025}
}
\bibliographystyle{icml2026}

\newpage
\appendix
\onecolumn

\section*{Overview of the Appendix}
The \textbf{appendix} is organized into the following sections:

\begin{itemize}
    \item \textbf{\cref{appendix:sec:notations}} explains all notations used throughout the paper.  
    \item \textbf{\cref{appendix:sec:detailed_proof}} presents detailed proofs for the mean and covariance under different types of transformation perturbations.  
    \item \textbf{\cref{appendix:sec:uncertainty_implement_details}} lists the details of the transformation perturbation and the statistics of our approach (e.g., runtime and memory consumption).
    \item \textbf{\cref{appendix:sec:experiment_settings}} describes the metrics, datasets, and backbone registration networks employed in the experiments.  
    \item \textbf{\cref{appendix:sec:dataset_level_evaluation}} details the experimental setup used in \cref{sec:dataset_level_evaluation}, including parameter choices, metric definitions, and figure generation. This section also discusses why the uncertainty map correlates with the error map and extends the dataset-level experiments to additional anatomical structures (e.g., cardiac).  
    \item \textbf{\cref{appendix:sec:exp_single_case}} provides single-case analyses on randomly sampled brain, abdomen, and cardiac images to demonstrate the generalization of our observations. 
    \item \textbf{\cref{appendix:sec:exp_case_study}} gives an extended discussion of the experiments and results in \cref{sec:exp_case_study}, and includes further case studies on BraTS-Reg and Learn2Reg ThoraxCBCT to illustrate the generalizability of our findings.  
\end{itemize}

\section{Appendix: Notations}\label{appendix:sec:notations}
\begin{table}[h!]
\centering
\renewcommand{\arraystretch}{1.2}
\begin{tabular}{p{0.18\linewidth} p{0.72\linewidth}}
\hline
\textbf{Symbol} & \textbf{Meaning} \\
\hline
$ I^A, I^B $ & Source and target images \\
$ \Omega $ & Image domain \\
$ \mathbb{I} $ & Identity transformation \\
$ \tau \in \text{Diff}(\Omega) $ & Random perturbation sampled from a diffeomorphic transformation family \\
$ \phi^{AB} $ & True transformation mapping points from $\Omega^B$ to $\Omega^A$ \\
$ \hat{\phi}^{AB} $ & Transformation predicted by the registration network $f_\theta(\cdot,\cdot)$ \\
$ \epsilon(y) $ & Registration error at voxel $y$ \\
$ \mu_\epsilon(\tau; y) $ & Mean of the error under perturbation $\tau$ \\
$ \Sigma_\epsilon(\tau; y) $ & Covariance of the error under perturbation $\tau$ \\
$ g_\tau(y) $ & Composed prediction under perturbation, $g_\tau(y) = \tau \circ \hat{\phi}^{A'B}(y)$ \\
$ J_\tau(y) $ & Jacobian of the perturbation $\tau$ at location $y$ \\
$ \mu(y) $ & Mean of composed outputs over perturbations \\
$ S(y) $ & Sample covariance of composed outputs over perturbations \\
$ u(y) $ & Scalar uncertainty score, $u(y) = \sqrt{\mathrm{tr}\, S(y)}$ \\
mError & The Euclidean (L2) norm between the predicted and ground-truth transformed positions,
$\| \phi^{AB}(y) - \hat{\phi}^{AB}(y) \|_2 $ with $y\in\Omega^B$, averaged over the ROI. \\
AURC & Area under risk–coverage curve, quantifies uncertainty–error trade-off \\
\hline
\end{tabular}
\caption{Summary of notations used throughout the paper.}
\end{table}

\section{Appendix: Detailed Proofs of Lemmas}\label{appendix:sec:detailed_proof}

\textbf{Preliminaries.}
Recall the perturbed pair and composed output:
\begin{equation*}
\hat{\phi}^{A'B}(y)=\big(\tau^{-1}\circ \phi^{AB}\big)(y)+\epsilon\tau(y),
\qquad
g_\tau(y)\;=\;\tau\circ\hat{\phi}^{A'B}(y)\
\;=\;\tau\!\left(\,(\tau^{-1}\!\circ \phi^{AB})(y)+\epsilon\tau(y)\,\right),
\end{equation*}
with $\epsilon_\tau(y)\sim \mathcal{N}\!\big(\mu_{\!\epsilon}(\tau;y),\Sigma_{\!\epsilon}(\tau;y)\big)
$.
Let $v_\tau(y):=(\tau^{-1}\!\circ \phi^{AB})(y)$ and denote the Jacobian of $\tau$ at $v(y)$ by $J_\tau(y):=D\tau|_{v(y)}$.
\textit{Definitions as in Sec.~3.2–3.3.} \,{\small(See main text for \cref{equ:purturbed_registration}, \cref{equ:composed_transformation} and the error model.)}
\footnote{We follow the composition and error setup in \cref{sec:composition_definision}–\cref{sec:theory_analysis}: \cref{equ:purturbed_registration} and \cref{equ:composed_transformation} define
$\hat{\phi}^{A'B}$ and $g_\tau$, and the error model for $\epsilon\tau$.}
\label{app:prelim}

\subsection*{Proof of \cref{lemma:uncertainty_proxy}} 
For completeness, we restate \cref{lemma:uncertainty_proxy} here:
\begin{quote}
\textbf{\cref{lemma:uncertainty_proxy}.} Let the perturbed output be
$g_{\tau}(y)
\;=\;\tau\!\left(\,(\tau^{-1}\!\circ \phi^{AB})(y)+\epsilon_\tau(y)\,\right)$, with
$\epsilon_\tau(y)\sim\mathcal N\!\bigl(\mu_{\!\epsilon}(\tau;y),
                                        \Sigma_{\!\epsilon}(\tau;y)\bigr)$. Without assuming any independence between $\tau$ and
$\epsilon_\tau$ we have
\begin{align}
\mathbb{E}_{\tau}\bigl[g_{\tau}(y)\bigr]
    &= \phi^{AB}(y)
       \;+\;
       \mathbb{E}_{\tau}\!\bigl[
           J_{\tau}(y)\,\mu_{\!\epsilon}(\tau;y)
         \bigr],
\\[4pt]
\operatorname{Cov}_{\tau}\!\bigl[g_{\tau}(y)\bigr]
    &= \mathbb{E}_{\tau}\!\bigl[
         J_{\tau}(y)\,
         \Sigma_{\!\epsilon}(\tau;y)\,
         J_{\tau}^{\!\top}(y)
       \bigr]
       \;+\;
       \operatorname{Cov}_{\tau}\!\bigl[
         J_{\tau}(y)\,\mu_{\!\epsilon}(\tau;y)
       \bigr].
\end{align}
\end{quote}

\begin{proof}
Let $v_\tau(y):=(\tau^{-1}\!\circ \phi^{AB})(y)$, so $\tau(v_\tau(y))=\phi^{AB}(y)$.
Apply a first-order Taylor expansion of $\tau$ at $v_\tau(y)$ to the argument $v_\tau(y)+\epsilon_\tau(y)$,
\begin{equation}
    \tau\!\big(v_\tau(y)+\epsilon_\tau(y)\big)
=\tau\!\big(v_\tau(y)\big)+J_\tau(y)\,\epsilon_\tau(y)+R_\tau(y),
\end{equation}
where the remainder $R_\tau(y)=o(\|\epsilon_\tau(y)\|)$ is higher order in the perturbation.
Because $\tau(v_\tau(y))=\phi^{AB}(y)$, the offset simplifies and we obtain (to first order)
\begin{equation}
g_\tau(y)\;=\;\phi^{AB}(y)\;+\;J_\tau(y)\,\epsilon_\tau(y)\;+\;R_\tau(y).
\end{equation}
Conditioning on $\tau$ and using the error model moments,
$\mathbb{E}[\,\epsilon_\tau(y)\!\mid\!\tau\,]=\mu_{\!\epsilon}(\tau;y)$,
$\operatorname{Cov}[\,\epsilon_\tau(y)\!\mid\!\tau\,]=\Sigma_{\!\epsilon}(\tau;y)$,
we get
\begin{equation}
    \mathbb{E}[\,g_\tau(y)\!\mid\!\tau\,]=\phi^{AB}(y)+J_\tau(y)\,\mu_{\!\epsilon}(\tau;y),
\quad
\operatorname{Cov}[\,g_\tau(y)\!\mid\!\tau\,]=J_\tau(y)\,\Sigma_{\!\epsilon}(\tau;y)\,J_\tau(y)^\top.
\end{equation}
Apply the laws of total expectation and total covariance:
\begin{equation}
\mathbb{E}_\tau[g_\tau(y)]=\mathbb{E}_\tau\big[\mathbb{E}[g_\tau(y)\!\mid\!\tau]\big],
\quad
\operatorname{Cov}_\tau[g_\tau(y)]
=\mathbb{E}_\tau\big[\operatorname{Cov}(g_\tau(y)\!\mid\!\tau)\big]
+\operatorname{Cov}_\tau\big(\mathbb{E}[g_\tau(y)\!\mid\!\tau]\big),
\end{equation}
which yield the stated formulas. Higher-order terms are neglected in the first-order approximation.
\end{proof}

\subsection*{Proof of \cref{lemma:affine_error}} 
For completeness, we restate \cref{lemma:affine_error} here:
\begin{quote}
\textbf{\cref{lemma:affine_error}.} 
Let $ \tau(z) = A z + b $ be a random affine transformation with $ A \in \mathbb{R}^{d \times d} $, $ b \in \mathbb{R}^d $. Since the translation part cancels, $g_{\tau}(y)=\phi^{AB}(y)+A\epsilon_\tau(y)$. Then:
\begin{align}
\mathbb{E}_{A,b}\!\bigl[g_{A,b}(y)\bigr]
  &= \phi^{AB}(y)\;+\;\mathbb{E}_{A}\!\bigl[A\,\mu_{\!\epsilon}(A;y)\bigr],\\[4pt]
\operatorname{Var}_{A,b}\!\bigl[g_{A,b}(y)\bigr]
  &= \mathbb{E}_{A}\!\bigl[A\,\Sigma_{\!\epsilon}(A;y)\,A^{\!\top}\bigr]
     \;+\;\operatorname{Var}_{A}\!\bigl[A\,\mu_{\!\epsilon}(A;y)\bigr].
\end{align}
\end{quote}

\begin{proof}
Write $v_\tau(y)=(\tau^{-1}\!\circ \phi^{AB})(y)$. Since $\tau$ is affine,
\begin{equation}
g_\tau(y)=\tau\!\big(v_\tau(y)+\epsilon_\tau(y)\big)
= A\big(v_\tau(y)+\epsilon_\tau(y)\big)+b
= \underbrace{(A v_\tau(y)+b)}_{=\;\tau(v_\tau(y))=\phi^{AB}(y)} \;+\; A\,\epsilon_\tau(y).
\end{equation}
Thus $g_\tau(y)=\phi^{AB}(y)+A\,\epsilon_\tau(y)$ holds exactly (no linearization).
Conditioning on $A$ (the distribution of $b$ cancels), we obtain
$\mathbb{E}[g_\tau(y)\!\mid\!A]=\phi^{AB}(y)+A\,\mu_{\!\epsilon}(A;y)$ and
$\operatorname{Cov}[g_\tau(y)\!\mid\!A]=A\,\Sigma_{\!\epsilon}(A;y)\,A^\top$.
Unconditioning via total expectation/covariance gives the result.
\end{proof}

\subsection*{Proof of Corollary~\ref{lemma:translation}}
For completeness, we restate Corollary~\ref{lemma:translation} here:
\begin{quote}
\textbf{Corollary~\ref{lemma:translation}.}
Let $\tau(z)=z+t$ with random $t$. Then $g_t(y)=\phi^{AB}(y)+\epsilon_t(y)$, and
\begin{align}
\mathbb{E}_{t}[g_t(y)]
 &= \phi^{AB}(y)+\mathbb{E}_{t}\!\bigl[\mu_{\!\epsilon}(t;y)\bigr],\\[4pt]
\operatorname{Var}_{t}[g_t(y)]
 &= \mathbb{E}_{t}\!\bigl[\Sigma_{\!\epsilon}(t;y)\bigr]
    +\operatorname{Var}_{t}\!\bigl[\mu_{\!\epsilon}(t;y)\bigr].
\end{align}
\end{quote}

\begin{proof}
For translations, $\tau^{-1}(z)=z-t$ and $\tau(v)=v+t$. Hence
\begin{equation}
g_t(y)=\tau\!\Big((\tau^{-1}\!\circ \phi^{AB})(y)+\epsilon_t(y)\Big)
=\big(\phi^{AB}(y)-t+\epsilon_t(y)\big)+t
=\phi^{AB}(y)+\epsilon_t(y),
\end{equation}
exactly. Conditioning on $t$ yields
$\mathbb{E}[g_t(y)\!\mid\!t]=\phi^{AB}(y)+\mu_{\!\epsilon}(t;y)$ and
$\operatorname{Var}[g_t(y)\!\mid\!t]=\Sigma_{\!\epsilon}(t;y)$.
Apply total expectation/variance to conclude.
\end{proof}

\textbf{Accuracy of the approximations.}
The proofs for affine and translation perturbations are exact (no Taylor remainder).
For general diffeomorphisms, \cref{lemma:uncertainty_proxy} follows from a first-order expansion at $v_\tau(y)$; higher-order terms in \(\epsilon_\tau(y)\) vanish as the perturbation scale decreases,
and the result is accurate in the small-perturbation regime customarily used in practice.

\textbf{Why do we choose the additive error model instead of the compositional error model in \cref{equ:error_model}?} Registration error can be formulated compositionally,
$\hat{\phi}^{AB} \circ (\phi^{AB})^{-1},$
or additively,
$\hat{\phi}^{AB} - \phi^{AB}.$ Both are mathematically valid: the compositional model reflects the geometry of diffeomorphic transformations, while an additive model reflects the standard regression view of predicting continuous-valued deformation fields. Our goal for the analysis is to rewrite the variance w.r.t. the error so that we can explain why the variance and error correlates and further interpret the uncertainty map. 
In practice, however, registration error is almost always computed additively (e.g., Euclidean distance between warped and target landmarks), which corresponds to $\hat{\phi}^{AB}(\cdot) - \phi^{AB}(\cdot)$. To keep the theoretical analysis consistent with the evaluation used in practice, we therefore adopt the additive error model.

\section{Appendix: Details of the Experiment Settings}\label{appendix:sec:details_of_experiments}
\subsection{Implementation of the transformation perturbation}\label{appendix:sec:uncertainty_implement_details}
The transformations for the perturbation are sampled from the following distribution.
\begin{itemize}
    \item Translation Perturbation: $t\sim\mathcal U(-t,t)$ with $t=1\%$ of image shape;
    \item Shear Perturbation: shear factor sampled from $\mathcal U(-0.02, 0.02)$;
    \item Scale Perturbation: scale factor sampled from $\mathcal U(0.9,1.1)$;
    \item B-spline transformation Perturbation (Deform): with grids distribution 10 pixels apart and displacement randomly sampled from $U(-12.5 px,12.5 px)$.
\end{itemize}
We sample $N=50$ transformations for each perturbation type and compute the variance across all the experiments.

\textbf{Approach Statistics.} We report the memory usage and runtime overhead introduced by our uncertainty estimation to quantify its inference-time cost. The measured runtime includes three components: loading the registration network, a one-time registration inference, and the computation of the uncertainty map with $N=50$ perturbations (as described in \cref{alg:spv}). For comparison, the baseline backbone inference time includes only model loading and a single registration pass, averaged over 50 runs to reduce variance. All experiments were conducted on an RTX 4500 Ada GPU. The results are summarized in \cref{appendix:tab:method_statistics}. Note that perturbation transformations are generated on the CPU, which could be further optimized to reduce runtime.

\begin{table}[ht]
    \centering
    \begin{tabular}{lccccc}\toprule
 & & \multicolumn{4}{c}{Uncertainty Estimation}\\ \cmidrule{3-6}
         &  uniGradICON&  Transformation&  Shear&  Scale& Deform\\ \midrule
         Runtime (s)& 0.20 & 12.09 & 12.17 & 12.13 & 27.72 \\
         GPU Memory (MB)& 2363 & 2630 & 2630 & 2630 & 2713 \\ \bottomrule
    \end{tabular}
    \caption{Runtime and peak memory of the backbone model uniGradICON and our proposed uncertainty estimation approach.}
    \label{appendix:tab:method_statistics}
\end{table}

\subsection{Datasets, backbones, and evaluation metrics}\label{appendix:sec:experiment_settings}
\textbf{Metrics.} We use three evaluation metrics: (1) Pearson correlation, (2) Spearman correlation, and (3) the area under the risk-coverage curve (AURC). The Pearson and Spearman correlations are computed between the uncertainty map and the voxel-wise registration error. These correlations are then averaged within each ROI and across the dataset to obtain instance-level and dataset-level correlation scores. The AURC is a scalar metric that quantifies the effectiveness of an uncertainty (or confidence) measure in identifying reliable predictions. It is computed from the risk–coverage curve, where coverage denotes the fraction of predictions retained after discarding the least confident ones, and risk denotes the corresponding prediction error on the retained set. By integrating risk over all coverage levels, the AURC summarizes how well uncertainty scores allow a trade-off between accuracy and coverage. For each voxel, the prediction error is computed as the Euclidean (L2) norm between the predicted and ground-truth transformed positions,
$\| \phi^{AB}(y) - \hat{\phi}^{AB}(y) \|_2 $ with $y\in\Omega^B$.

\textbf{Datasets.} In the experiments, we prepare four datasets that cover brain MRI, cardiac MRI, and abdomen CT. These datasets are used for quantitative evaluations. In addition, we conduct case studies on Brats-Reg \citep{baheti2021brain} and Learn2Reg ThoraxCBCT \citep{hugo2016data, hugo2017longitudinal} datasets. 

\emph{Curated Brain MRI.} We collect a brain MRI dataset by randomly sampling 20 MRI images from 11 brain MRI repositories \ (ABIDE \citep{di2014autism}, ADHD200 \citep{brown2012adhd}, ADNI \citep{jack2008alzheimer,weiner2017alzheimer}, AIBL \citep{fowler2021fifteen}, FreeSurfer \citep{fischl2002whole}, COBRE \citep{mayer2013functional}, Chinese-HCP \citep{vogt2023chinese}, HCP \citep{van2012human}, ISBI2015 \citep{carass2017longitudinal}, MCIC \citep{gollub2013mcic}, OASIS3 \citep{lamontagne2019oasis}), leading to a 220 MRI volumes dataset. This curated dataset consists of highly heterogeneous populations, including healthy subjects (HCP, Chinese-HCP), dementia (OASIS3, ADNI3, AIBL, FreeSurfer), Autism Spectrum Disorder (ABIDE), Attention Deficit Hyperactivity Disorder (ADHD), Schizophrenia (COBRE, MCIC), and Multiple Sclerosis (ISBI2015). We use the whole dataset in the evaluation with a real transformation experiment, and use 110 images for the simulated transformation experiment.

\emph{ACDC Cardiac MRI.} We use the training set from the ACDC dataset \citep{bernard2018deep}, which contains 100 pairs of cardiac MRI that are acquired at the end of the diastolic (ED) and systolic (ES) phase of the same subject.

\emph{Learn2Reg Abdomen CT.} We use the training set of the Abdomen CT dataset \citep{xu2016evaluation} released by the Learn2Reg registration challenge, which contains 30 abdomen CT images. For the synthetic transformation experiment, we use the 30 images with synthetic transformations. In the real transformation experiment, we use the 30 images as the source image and randomly sample 30 images from the same list of images as the target image, composing the 30 pairs of images.

\emph{IXI Brain MRI.} We use the IXI Brain MRI dataset following the preprocessing pipeline provided by TransMorph~\citep{chen2022transmorph}, and we adopt the same test split as in TransMorph. The processed dataset contains one atlas and 115 brain MRIs.

\textbf{Backbone registration models.} We use two deterministic registration foundation models and one probabilistic registration model as the backbones. We obtain the models, including their weights, from their official GitHub repositories.

\emph{uniGradICON \citep{tian2024unigradicon}.} It is a medical image registration foundation model that is trained across a curated heterogeneous dataset covering several anatomical structures. It demonstrates excellent generalization to both in-distribution and out-of-distribution registration tasks.

\emph{SynthMorph \citep{hoffmann2021synthmorph}.} It is a modality-agnostic brain registration network that is trained purely on synthetic brain MRI and demonstrates great generalization and robustness for brain image registration. We also test this model on the cardiac and abdominal images. We note that SynthMorph provides both an affine model and a nonlinear model, and we use the nonlinear model in our experiments. We do not aim to measure the registration accuracy but the validity of the uncertainty estimation, namely, the correlation between the registration error and the estimated confidence score. 

\emph{TransMorph \citep{chen2022transmorph}}. It is a transformer-based registration network that provides variant pre-trained models, including a probabilistic registration model trained with MC-dropout.

\subsection{Dataset-level Quantitative Analysis}\label{appendix:sec:dataset_level_evaluation}
In the main manuscript (\cref{sec:dataset_level_evaluation}), we provide a concise description of the experimental setup, present selected results, and summarize the main takeaways. In this appendix, we provide details of the experiment setting, report the complete results (including the ACDC cardiac MRI dataset), and offer an extended discussion.

\textbf{Experiment Setting.} To quantitatively evaluate the correlation between the registration error map and the uncertainty map estimated by our method, we simulate three types of deformation and apply the randomly simulated deformation to warp the image to get the paired image with ground truth deformation. We test the uncertainty estimation under the following three simulated ground truth deformations: 
\begin{itemize}
    \item Translation $t\sim\mathcal U(-t,t)$ with $t=10\%$ of image shape;
    \item Affine $\tau(z)=Az+b$ with translation uniformaly sampled in $\mathcal U(-t,t)$ with $t=10\%$ of image shape, shear factor sampled from $\mathcal U(-0.1, 0.1)$, and scale factor sampled from $\mathcal U(0.8,1.2)$;
    \item Composition of two elastic B-spline transformations with grids distribution 10 pixels apart and displacement randomly sampled from $U(-12.5 px,12.5 px)$\footnote{12.5 pixel is the largest displacement one can for a B-spline transformation that has grids 10 pixels apart and a shape of $175\times175\times175$, which is the default shape we use in our experiments.}.
\end{itemize}
 We use a composition of two elastic deformations in (iii) to mimic an extremely large deformation scenario (as shown in \cref{fig:sim_transform_visulization}).

While simulated transformations provide controlled conditions for evaluating registration uncertainty, they do not fully reflect the challenges in real registration scenarios. We design an experiment setting that approximates real-world conditions while providing access to ground-truth transformations. Specifically, we use a conventional registration algorithm (ANTs) \citep{avants2008symmetric} to estimate the transformations between real image pairs with affine followed by non-rigid registration. These estimated transformations are then applied to warp the source image, producing a new target image that shares the same transformation. Although this is a workaround, it ensures that the ground-truth transformations are derived from real registration problems, thereby closely mimicking practical registration scenarios while preserving access to reference transformations. We follow the same experimental design as in the simulated transformation study, applying different perturbation types (translation, affine, and nonrigid deformation) to evaluate both uniGradICON and SynthMorph. 

\newpage

\begin{figure}[h]
    \centering
    \begin{subfigure}[b]{\textwidth}
    \includegraphics[width=\linewidth]{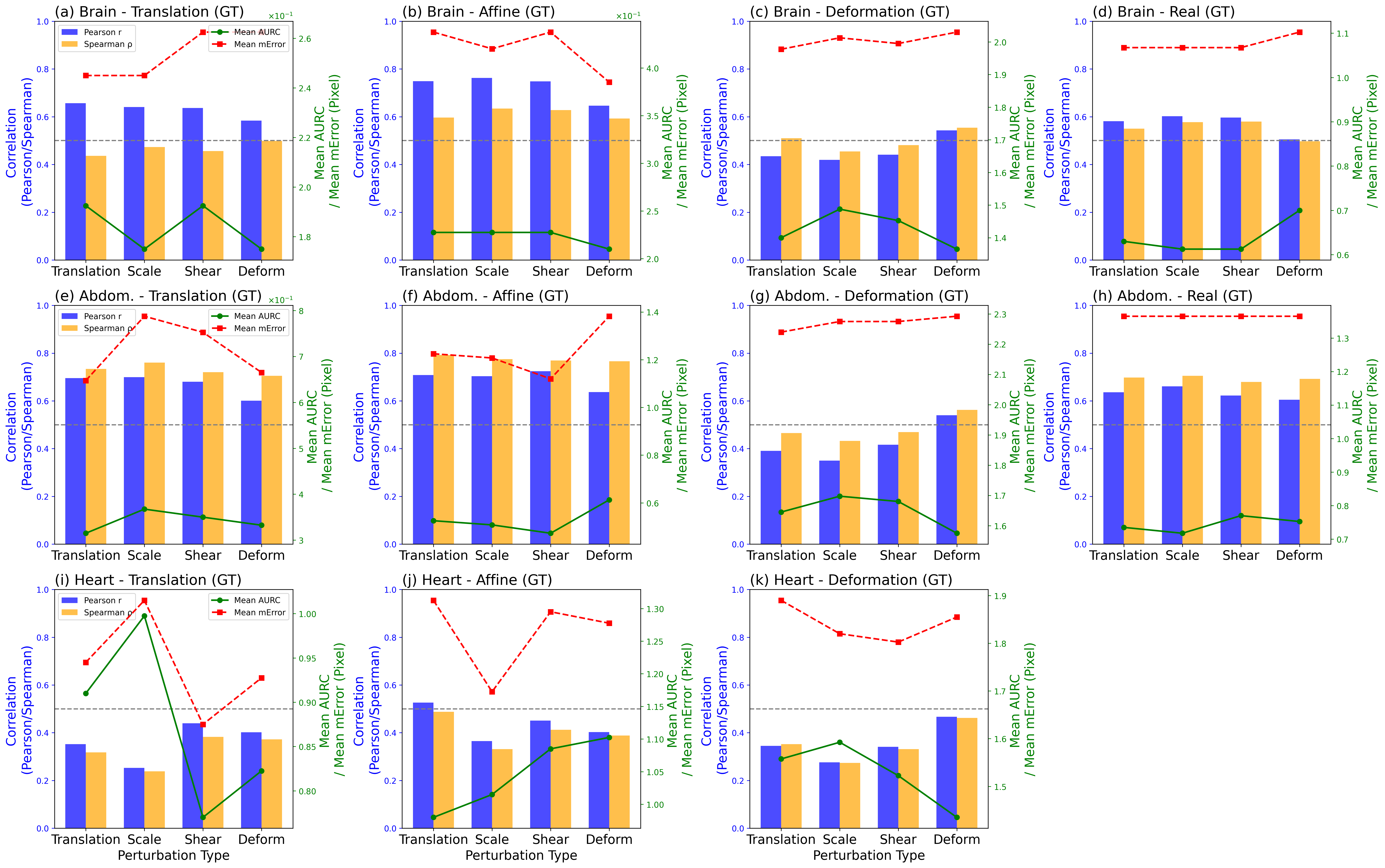}
    \caption{Evaluation of confidence score with uniGradICON.}
    \label{appendix:fig:correlation_full_unigradicon}
    \end{subfigure}

    \medskip %
    \begin{subfigure}[b]{\textwidth}
      \includegraphics[width=\linewidth]{figures/sim_correlation_comparison_synthmorph.png}
      \caption{Evaluation of confidence score with SynthMorph.}
      \label{appendix:fig:correlation_full_synthmorph}
      \end{subfigure}
\caption{Correlation between the estimated confidence score and the true registration error for a combination of deformation types, dataset types, and perturbation types.}
\label{appendix:fig:correlation_full}
\end{figure}

\cref{appendix:fig:correlation_full} presents additional results beyond \cref{fig:correlation_unigradicon} and \cref{fig:correlation_synthmorph}.
For \textbf{uniGradICON}, \cref{appendix:fig:correlation_full_unigradicon} reports additional results on the cardiac dataset (\cref{appendix:fig:correlation_full_unigradicon} (i)–(k)). Overall correlation levels are lower than those observed for the brain and abdominal datasets. We hypothesize that this reduction is related to the use of label maps of cardiac structures as the ROI mask. As illustrated in \cref{appendix:fig:exp_single_case_cardiac}, the resulting ROI is relatively small, leading to a smaller number of samples used to compute the correlation. As cardiac ROI contains fewer voxels, the resulting correlation estimates are noisier and less stable. Importantly, the correlation remains positive and moderate, indicating that the observed trend persists across tasks.

For \textbf{SynthMorph}, we further show the correlation between the uncertainty map and the registration error map under translation and affine ground truth transformation with brain MRI (\cref{appendix:fig:correlation_full_synthmorph} (a)-(b)). The correlations under translation and affine dropped quickly. This observation leads to the following analysis of why and when the uncertainty map is correlated with the registration error map.

\textbf{The intuition of why the variance relates to the error?}
With the general form of the error model, for a single unperturbed prediction of the registration network, the \emph{per-voxel registration error} decomposes into bias and variance
\[
\underbrace{\mathrm{{Error}}(y)}_{\text{total error}}
  \;=\;
  \mathbb{E}\!\bigl[\|\epsilon(y)\|^{2}\bigr]
  \;=\;
  \|\mu_{\!\epsilon}(y)\|^{2}
  \;+\;
  \underset{\text{shared term}}{\underbrace{\operatorname{tr}\Sigma_{\!\epsilon}(y)}}.
\]
By contrast, the perturbation variance derived in~\cref{eq:cov_general} is
\[
\underbrace{\Var_{\tau}\bigl[g_{\tau}(y)\bigr]}_{\text{uncertainty map}}
  \;=\;
  \underset{\text{shared term}}{\underbrace{\E_{\tau}\!\bigl[J_{\tau}\,
        \Sigma_{\!\epsilon}(\tau;y)\,J_{\tau}^{\!\top}\bigr]}}
  \;+\;
  \Var_{\tau}\!\bigl[
        J_{\tau}\,\mu_{\!\epsilon}(\tau;y)\bigr]\!.
\]

Both quantities contain the \emph{mean intrinsic spread}  
$\E_{\tau}[J_{\tau}\Sigma_{\!\epsilon}J_{\tau}^{\!\top}]$.
They differ in the other term:

\begin{center}
\begin{tabular}{r@{ }c@{ }l}
perturbation variance & $\longleftarrow$ &
$\Var_{\tau}\bigl[J_{\tau}\mu_{\!\epsilon}(\tau)\bigr]$
\quad (how \emph{variance the centers are} moving) \\[4pt]
total error            & $\longleftarrow$ &
$\|\mu_{\!\epsilon}\|^{2}$
\quad (how \emph{far the centre is} from zero).
\end{tabular}
\end{center}

When computing the correlation between the uncertainty map and the registration error, we will see the following results:
\begin{enumerate}[leftmargin=1.45em,itemsep=2pt,label=\alph*)]
\item \textbf{High correlation} occurs when the squared bias $||\mu_{\epsilon}(y)||^{2}$ and its perturbation-induced jitter $\Var_{\tau}\bigl[J_{\tau}\mu_{\!\epsilon}(\tau)\bigr]$ covary spatially. In this case, both maps increase in regions where either component grows.
\item \textbf{Low or negative correlation} can arise in two scenarios:  
\begin{enumerate}[leftmargin=1.8em,itemsep=2pt,label=(\roman*)]\label{appendix:sec:intuition_of_correlation_low_correlation}
\item when the network is \emph{more wrong than uncertain}, i.e., $||\mu_{\epsilon}(y)||^{2}$ dominates $\Var_{\tau}\bigl[J_{\tau}\mu_{\!\epsilon}(\tau)\bigr]$, such that large systematic errors are not reflected by the variance-based uncertainty;
\item when the network is highly uncertain about its prediction, producing a large $\Var_{\tau}\bigl[J_{\tau}\mu_{\!\epsilon}(\tau)\bigr]$ that does not spatially align with the bias term. 
\end{enumerate}
\end{enumerate}

It is important to note that low correlation does not always imply that the uncertainty map is uninformative. Case (i) reflects a true underestimation, where the uncertainty fails to capture systematic bias. In contrast, case (ii) provides a valid signal of model indecision: even if the registration happens to be accurate, the high uncertainty correctly reflects low confidence and should still draw attention.

Under this interpretation, we attribute the low correlation observed for \textbf{SynthMorph} on translation and affine ground truth transformations primarily to scenario (b)(i). SynthMorph’s inductive bias leads to relatively large systematic errors $||\mu_{\epsilon}(y)||^{2}$ in the presence of substantial global motion, which are not captured by variance-based uncertainty. This behavior is consistent with SynthMorph being designed for nonlinear registration and not for large linear transformations, as also reflected by its higher error range compared to uniGradICON (mean mError shown as the dotted \textcolor{red}{red line} in \cref{appendix:fig:correlation_full_unigradicon}-(a) and \cref{appendix:fig:correlation_full_synthmorph}-(a)). When the task falls within SynthMorph’s effective distribution (\cref{fig:correlation_synthmorph}(c)), the resulting correlations are consistent with those observed for uniGradICON.

In summary, beyond using the proposed variance to estimate uncertainty, in this experiment, we evaluate whether this uncertainty can act as a proxy for registration error. Empirically, we observe strong to moderate correlations when the registration task lies within the model’s capability. When the network is more wrong than uncertain, the uncertainty map no longer correlates with the registration error map. Importantly, this does not invalidate the uncertainty map itself: it still reflects ambiguity in establishing correspondences, but it is not a reliable calibration signal for registration error in this regime. This \textbf{limitation} is acceptable in practice, as nonlinear SynthMorph is not typically applied to scenarios involving large linear transformations. \textbf{In sum}, these results demonstrate that perturbation-based uncertainty provides a reliable proxy for registration error when model assumptions are satisfied.

\subsection{Single-Case Quantitative Analysis}\label{appendix:sec:exp_single_case}
To further probe the behavior of the proposed uncertainty map, we conduct a single-case analysis using randomly sampled subjects from the brain, abdominal, and cardiac datasets, following the settings in dataset-level analysis (\cref{appendix:sec:dataset_level_evaluation}). We evaluate the results qualitatively and quantitatively by visualizing uncertainty maps, error maps, and risk–coverage behavior, as well as reporting correlation and AURC metrics in \cref{fig:single_case_analysis_brain}.

\subsubsection{Single Case Analysis for Brain}

\begin{figure}[t]
\centering
\begin{subfigure}[b]{\textwidth}
    \includegraphics[width=\linewidth]{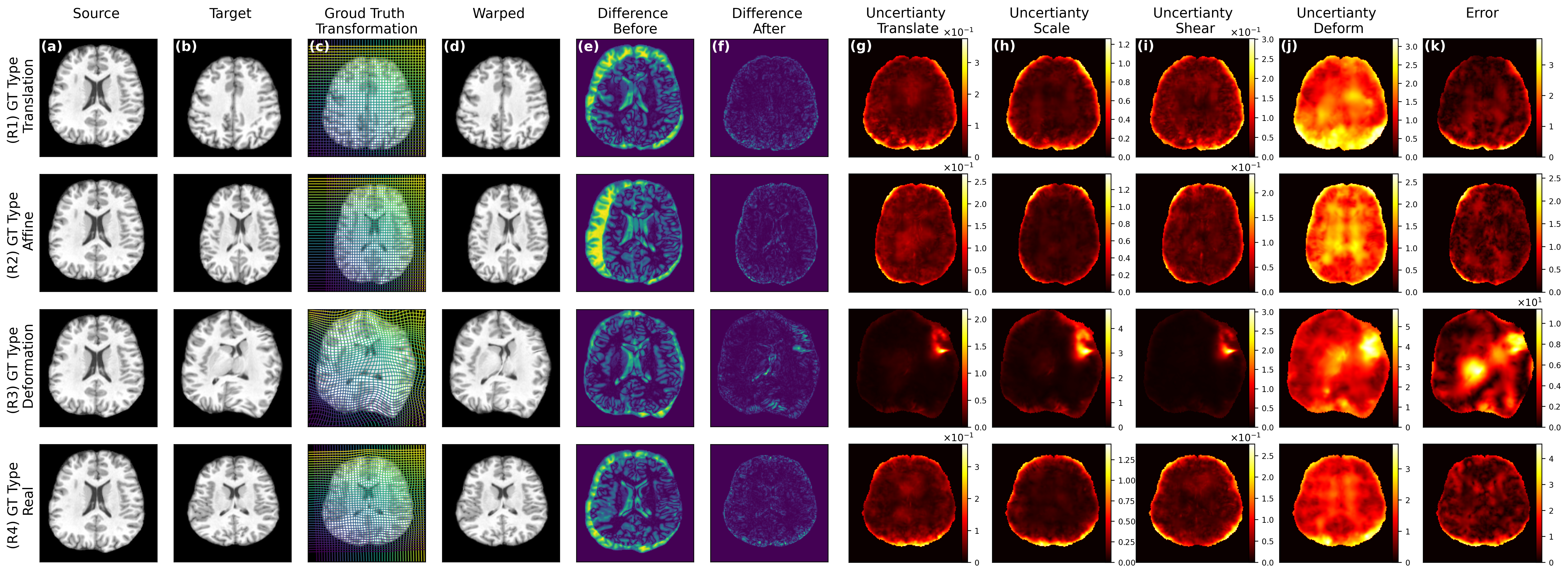}
    \caption{The qualitative results for one randomly sampled brain MRI case.}
    \label{fig:sim_transform_visulization}
\end{subfigure}

\medskip %
\begin{subfigure}[b]{\textwidth}
    \includegraphics[width=\linewidth]{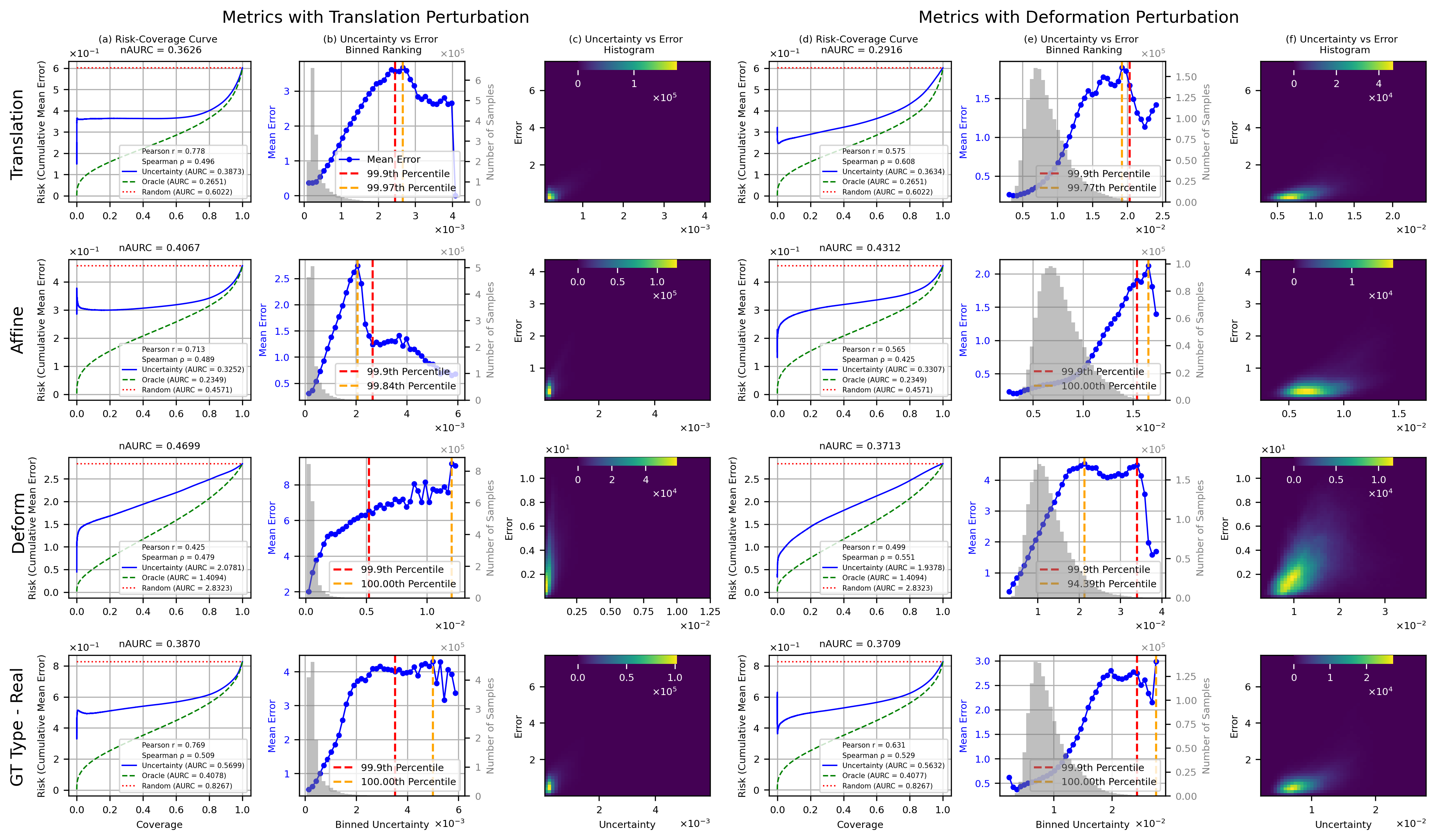}
    \caption{The quantitative results for one randomly sampled brain MRI case.}
    \label{fig:sim_transform_metrics}
\end{subfigure}
\caption{The qualitative and quantitative results of the uncertainty map measured of a randomly sampled brain MRI image for a combination of varying ground truth transformation types (GT), perturbation transformation types with uniGradICON.}
\label{fig:single_case_analysis_brain}
\end{figure}

\cref{fig:single_case_analysis_brain} shows qualitative and quantitative results for one randomly sampled brain MRI case. Under linear ground truths (translation and affine; \cref{fig:sim_transform_visulization} Row 1, Row 2), the uncertainty maps closely resemble the error maps ($k$-th column, denoted as C(k)), with the exception of the deformation-based uncertainty map (C(j)). In contrast, under a nonlinear ground truth (Row 3), the uncertainty map derived from nonlinear perturbations (C(j)) aligns more closely with the error than those derived from linear perturbations (C(g–i)), reflecting the interplay described in \cref{sec:dataset_level_evaluation}. Importantly, the error map under nonlinear ground truth (Row 3) exhibits more localized errors than those under linear ground truths (Row 1, Row 2). Nonlinear perturbations can reveal such local discrepancies, which translation-based perturbations fail to capture, underscoring their complementary strengths. The \textbf{quantitative results} over the full 3D volume (\cref{fig:sim_transform_metrics}) confirm these trends. Translation-based uncertainty achieves strong correlations with the true error ($r=0.71$–$0.78$, $\rho=0.49$–$0.50$) under translation, affine, and real ground truths, and moderate correlation ($r=0.43$, $\rho=0.48$) under deformation. To assess ranking ability, we compute normalized AURC, $\text{nAURC}=\frac{\text{Uncertainty}-\text{Oracle}}{\text{Random}-\text{Oracle}}$, which measures performance relative to random and oracle rankings. Lower values indicate better ranking. Across all ground-truth types, our method achieves low nAURC (0.29–0.47), demonstrating strong ranking capability. The second panel of each experiment (\cref{fig:sim_transform_metrics} C(b,e)) visualizes the mean error across bins, sorted by uncertainty, with bin sizes indicated in gray. Most errors are correctly prioritized (94.4th to 99.9th percentile), supported by the histogram (\cref{fig:sim_transform_metrics} C(c,f)). Overall, the results demonstrate that the proposed confidence score reliably highlights misalignments, achieves a strong correlation under global transformations, maintains a reasonable correlation under nonlinear deformations, and consistently separates high- and low-error regions. 

We include the result for another randomly sampled Brain MRI in \cref{appendix:fig:single_case_analysis_brain} to demonstrate the generalization of our discussion.

\begin{figure}[ht]
\centering
\begin{subfigure}[b]{\textwidth}
    \includegraphics[width=\linewidth]{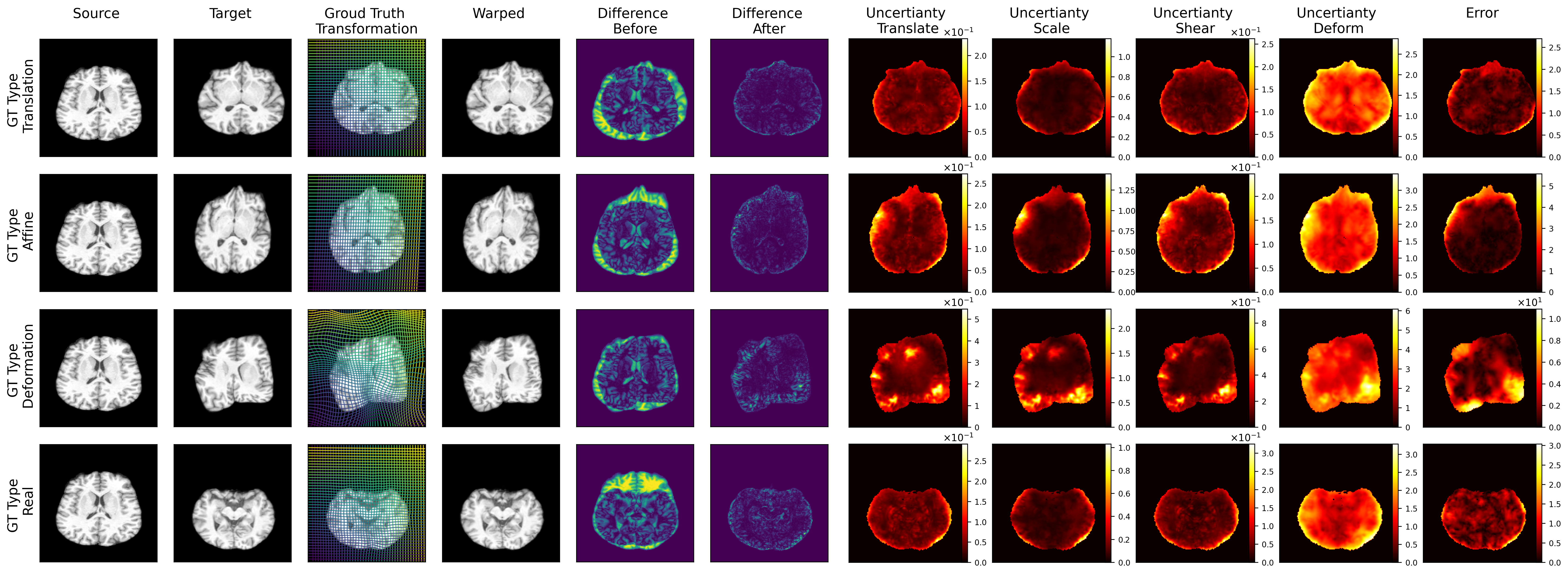}
    \caption{The visualization.}
    \label{appendix:fig:sim_transform_visulization}
\end{subfigure}

\medskip %
\begin{subfigure}[b]{\textwidth}
    \includegraphics[width=\linewidth]{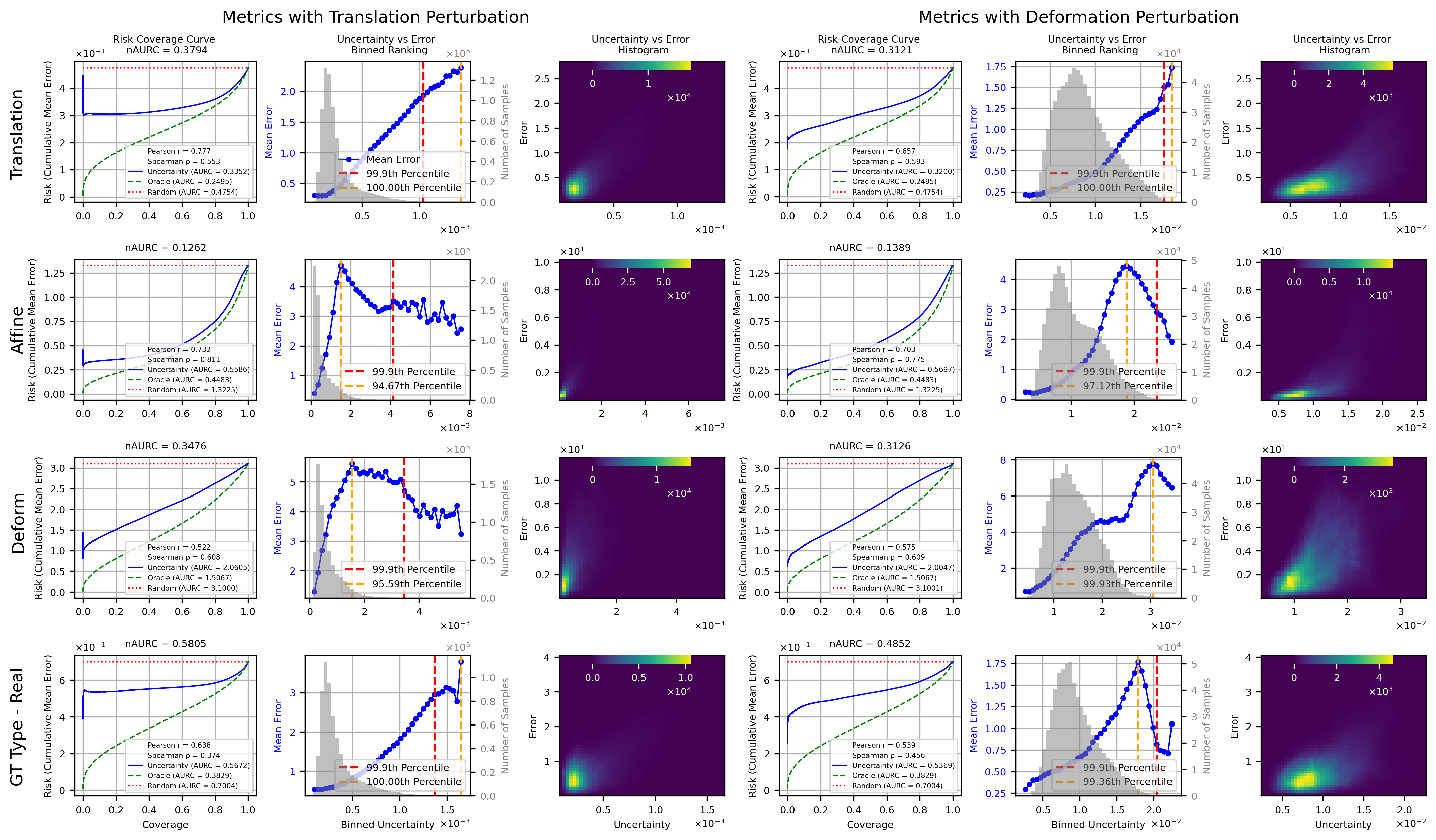}
    \caption{The qualitative results.}
    \label{appendix:fig:sim_transform_metrics}
\end{subfigure}
\caption{The qualitative and quantitative results of the uncertainty map measured of a randomly sampled brain MRI image from \textbf{COBRE} for a combination of varying ground truth transformation types (GT), perturbation transformation types with uniGradICON.}
\label{appendix:fig:single_case_analysis_brain}
\end{figure}

\clearpage
\subsubsection{Single Case Analysis for Abdomen}
We conduct the same analysis on one random abdomen CT image sampled from Learn2Reg Abdomen CT. The results are shown in \cref{appendix:fig:exp_single_case_abdomen_vis} and \cref{appendix:fig:exp_single_case_abdomen_metric}.

\begin{figure}
\centering
\begin{subfigure}[b]{\textwidth}
    \centering
    \includegraphics[width=\linewidth]{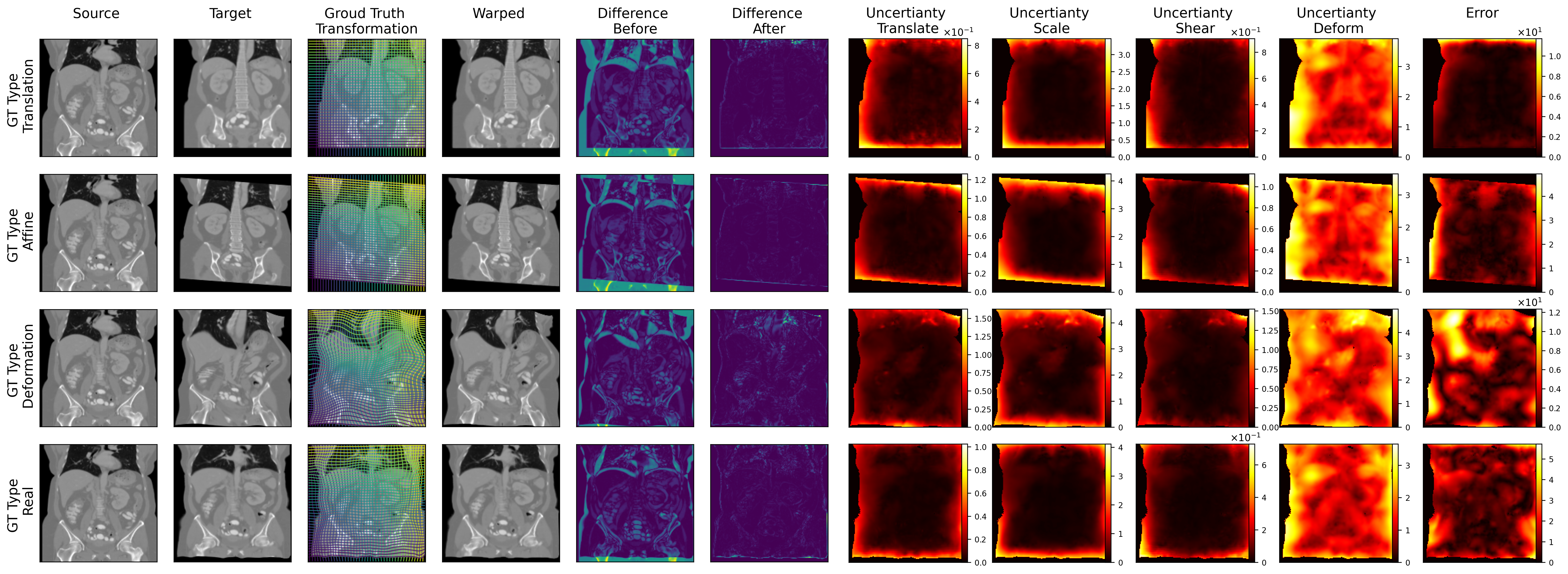}
    \caption{The visualization.}
    \label{appendix:fig:exp_single_case_abdomen_vis}
\end{subfigure}
\begin{subfigure}[b]{\textwidth}
    \centering
    \includegraphics[width=\linewidth]{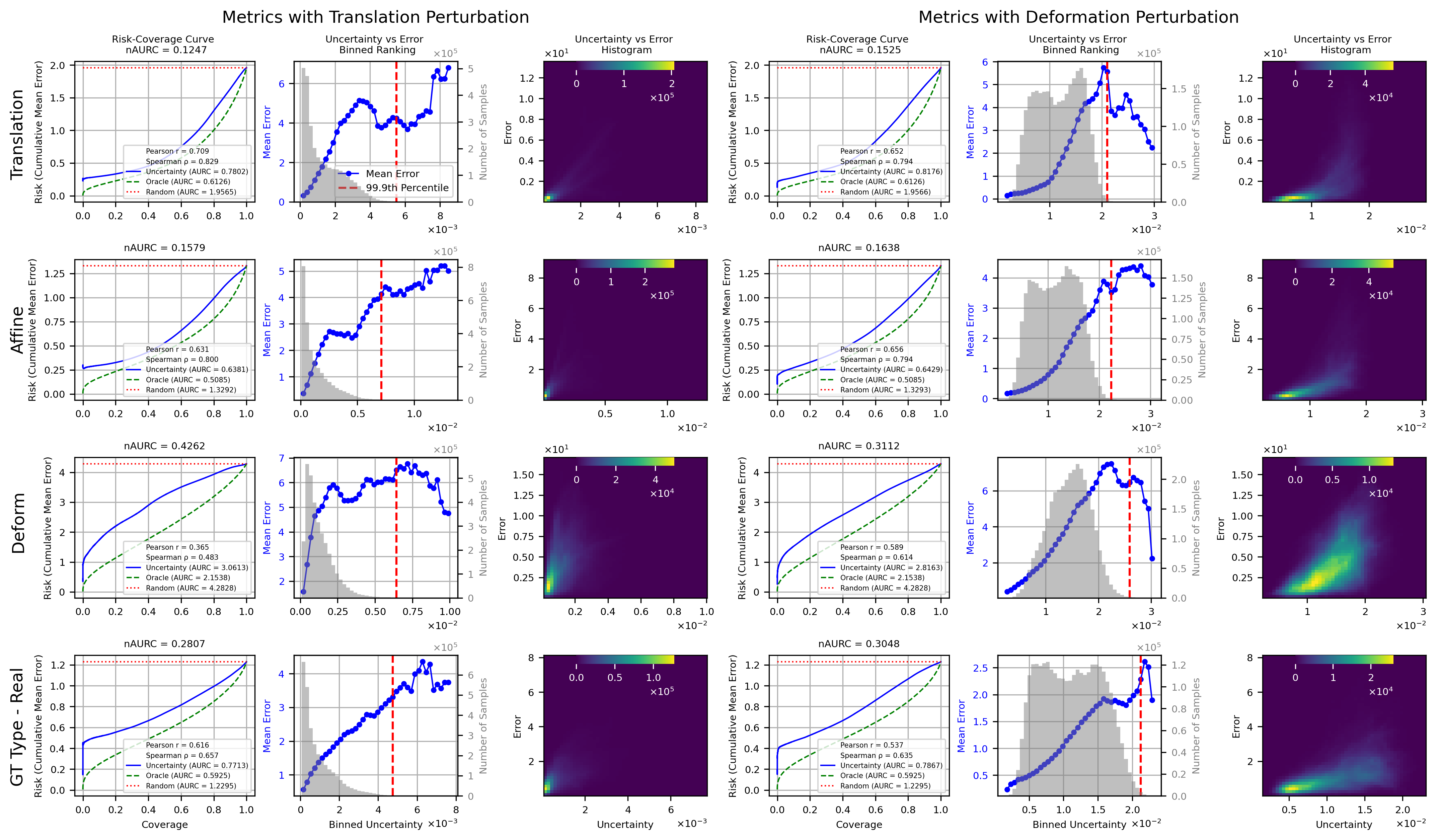}
    \caption{The quantitative results.}
    \label{appendix:fig:exp_single_case_abdomen_metric}
\end{subfigure}
\caption{The qualitative and quantitative results of the uncertainty map measured of a randomly sampled abdomen CT image for a combination of varying ground truth transformation types (GT), perturbation transformation types with uniGradICON.}
\label{appendix:fig:exp_single_case_abdomen}
\end{figure}

\textbf{Discussion.}
\cref{appendix:fig:exp_single_case_abdomen} shows qualitative and quantitative results for one randomly sampled abdomen CT case. The qualitative visualization (\cref{appendix:fig:exp_single_case_abdomen_vis}) demonstrates that the proposed uncertainty maps consistently highlight regions of misalignment. In addition, it also shows consistent observations as for the brain dataset:
\begin{itemize}
    \item Error maps vary across ground-truth types (last column in \cref{appendix:fig:exp_single_case_abdomen_vis}). The registration backbone tends to make localized registration errors when the ground truth transformation is nonlinear.
    \item Uncertainty maps vary across perturbation types under the same ground-truth transformation. As shown by the third row in \cref{appendix:fig:exp_single_case_abdomen_vis}, linear perturbations often fail to reveal the localized errors (columns 7-9). Nonlinear perturbations distort the geometry more strongly and are therefore more effective at exposing unstable regions.
\end{itemize} 

The quantitative evaluation (\cref{appendix:fig:exp_single_case_abdomen_metric}) aligns with these qualitative findings. Translation-based perturbations achieve higher correlations with the true error (Pearson $r \approx 0.63$, Spearman $\rho \approx 0.47$) and lower nAURC values when the ground truth transformation is translation or affine, which is consistent with the qualitative evaluation in~\cref{appendix:fig:exp_single_case_abdomen_vis}.
Moreover, for nonlinear ground truth transformations (third row in~\cref{appendix:fig:exp_single_case_abdomen_metric}), the binned error-vs-uncertainty curves of the translation perturbation flatten earlier than the deformation perturbation, indicating that many high-error voxels cannot be distinguished by uncertainty under linear perturbations, again showing that linear perturbations capture a different aspect of model instability.

\clearpage

\subsubsection{Single Case Analysis for Cardiac}

We conduct the same analysis on a randomly sampled cardiac MRI from the ACDC dataset. Results are presented in \cref{appendix:fig:exp_single_case_cardiac_vis} and \cref{appendix:fig:exp_single_case_cardiac_metric}. The ROIs, derived from segmentation maps of heart structures, are relatively small compared to the full image and contain limited texture variation. This lack of visual detail likely explains the elevated uncertainty observed in the maps, as the model faces inherent ambiguity in such low-texture regions. Consequently, high uncertainty arises even when registration error is limited, corresponding to scenario \ref{appendix:sec:intuition_of_correlation_low_correlation}-(ii) in the analysis of the proposed uncertainty maps and error maps, where uncertainty reflects low model confidence rather than direct error magnitude.

\begin{figure}[ht]
\centering
\begin{subfigure}[b]{\textwidth}
    \centering
    \includegraphics[width=\linewidth]{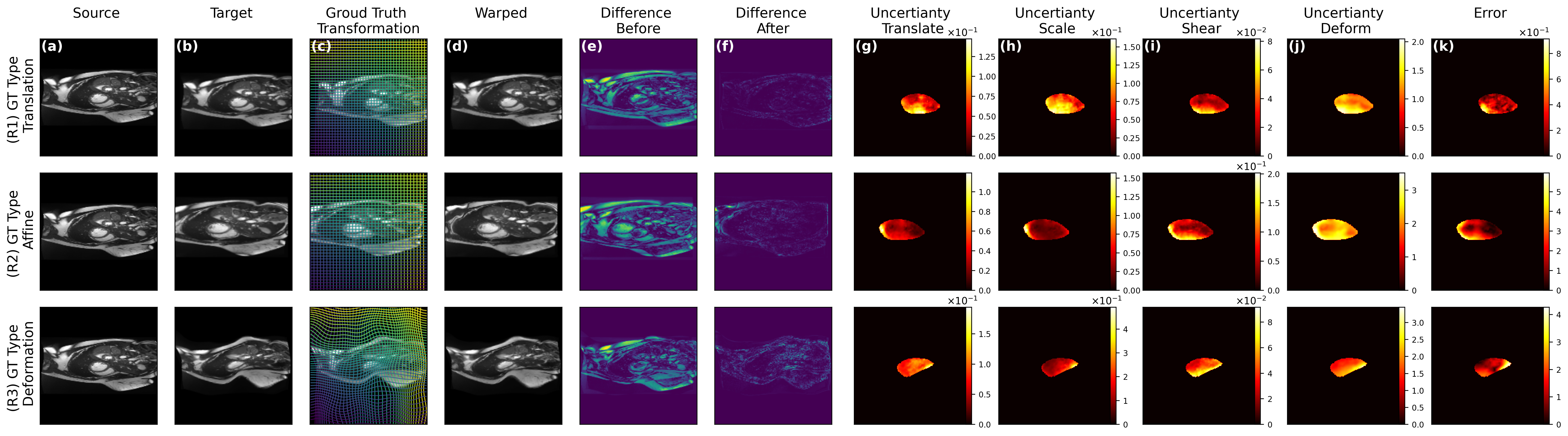}
    \caption{The visualization.}
    \label{appendix:fig:exp_single_case_cardiac_vis}
\end{subfigure}
\begin{subfigure}[b]{\textwidth}
    \centering
    \includegraphics[width=\linewidth]{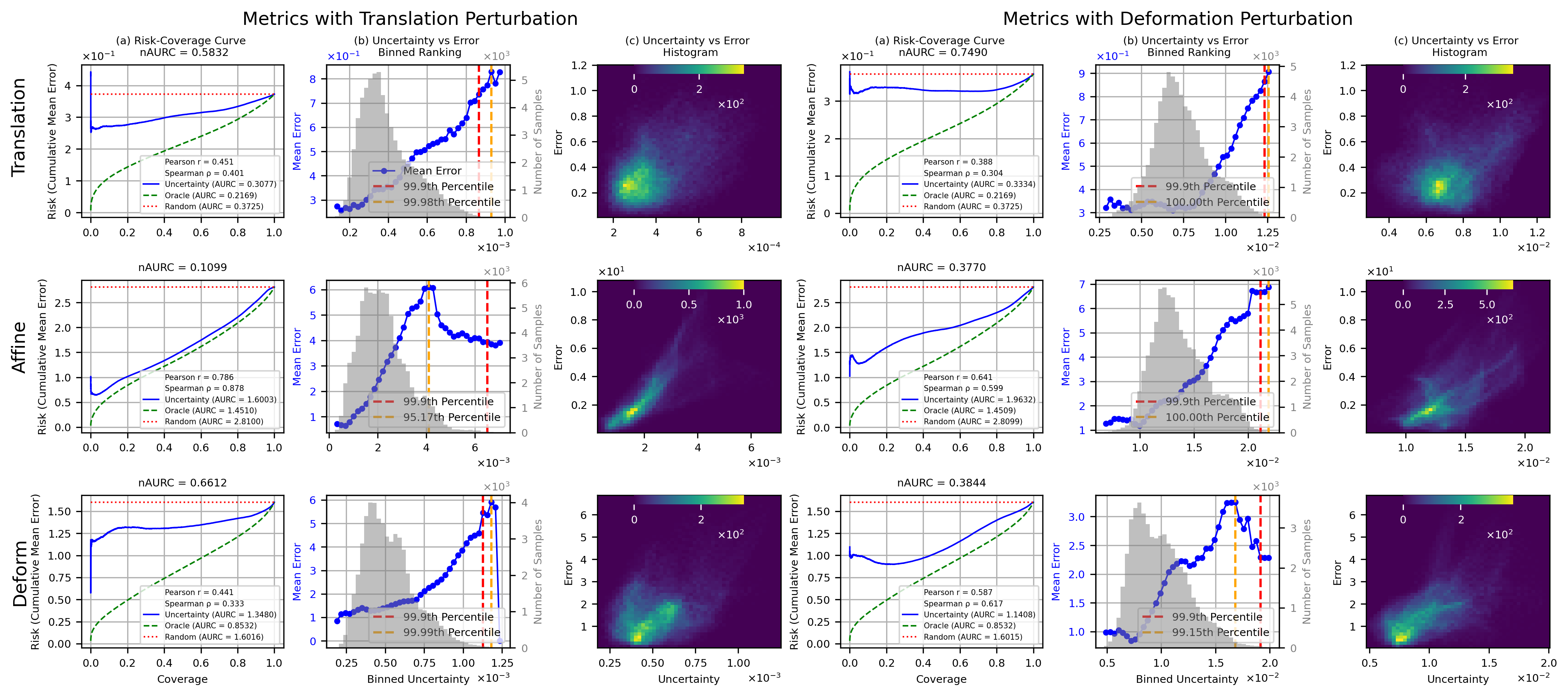}
    \caption{The quantitative results.}
    \label{appendix:fig:exp_single_case_cardiac_metric}
\end{subfigure}
\caption{The qualitative and quantitative results of the uncertainty map measured of a randomly sampled cardiac MRI image for a combination of varying ground truth transformation types (GT), perturbation transformation types with uniGradICON.}
\label{appendix:fig:exp_single_case_cardiac}
\end{figure}

\clearpage
\subsection{Case Study: Uncertainty Maps with Anatomical Inconsistencies}
\label{appendix:sec:exp_case_study}
In the main manuscript (\cref{sec:exp_case_study}), we describe two representative case studies to demonstrate how the proposed uncertainty maps highlight clinically relevant inconsistencies, such as tumor resection changes and field-of-view differences. In this appendix, we provide further details of the experimental design, including additional examples from the BraTS-Reg and Learn2Reg Thorax CBCT datasets, and offer a more extensive discussion of the observations. This experiment is run with uniGradICON.

\subsubsection{Case Study of the Pre-operative and Follow-up MRIs}
The first case study investigates whether the uncertainty map can identify inconsistent anatomical structures, such as pathological changes in the brain, before and after tumor resection. Beyond the example shown in the main manuscript (\cref{fig:exp_case_study_one_bratsreg}), we additionally sample five pairs of pre-operative and follow-up MRIs from the BraTS-Reg dataset \citep{baheti2021brain} and visualize the results in \cref{appendix:fig:exp_case_study_one_bratsreg}. Registrations and corresponding uncertainty maps are computed from T1-weighted images, while T2-weighted images are also included to better delineate abnormal regions. The results show that the uncertainty maps consistently highlight tumor-related inconsistencies in Cases 023, 030, 034, and 080, in line with the observations reported in \cref{sec:exp_case_study}. In contrast, Case 037 exhibits no prominent high-uncertainty regions, which can be attributed to the strong visual similarity between its source and target images. Taken together, these results confirm that the proposed uncertainty map generalizes beyond a single example and can reliably flag pathological inconsistencies across different subjects.

\begin{figure}[ht]
\centering
\begin{subfigure}[b]{\textwidth}
    \centering
    \includegraphics[width=\linewidth]{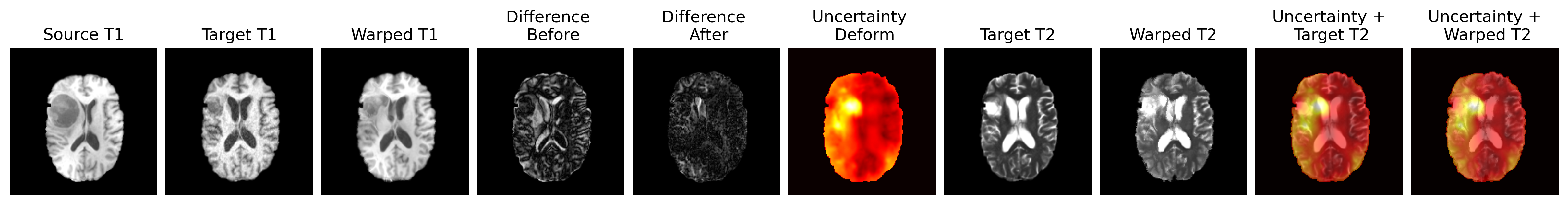}
    \caption{Case 023.}
    \label{appendix:fig:exp_case_study_one_bratsreg_023}
\end{subfigure}
\begin{subfigure}[b]{\textwidth}
    \centering
    \includegraphics[width=\linewidth]{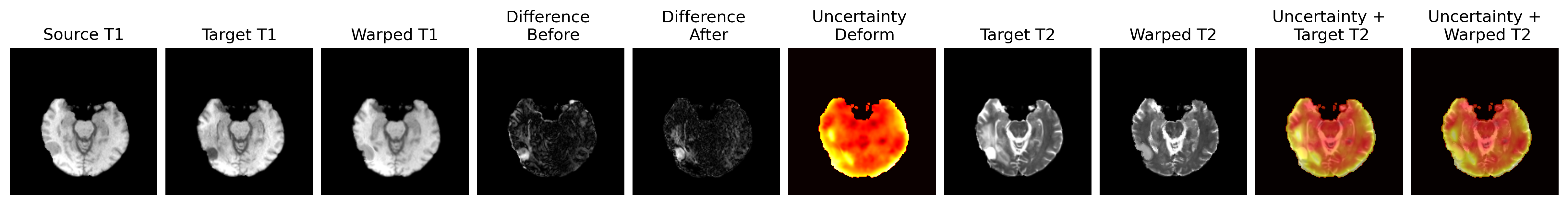}
    \caption{Case 030.}
    \label{appendix:fig:exp_case_study_one_bratsreg_030}
\end{subfigure}
\begin{subfigure}[b]{\textwidth}
    \centering
    \includegraphics[width=\linewidth]{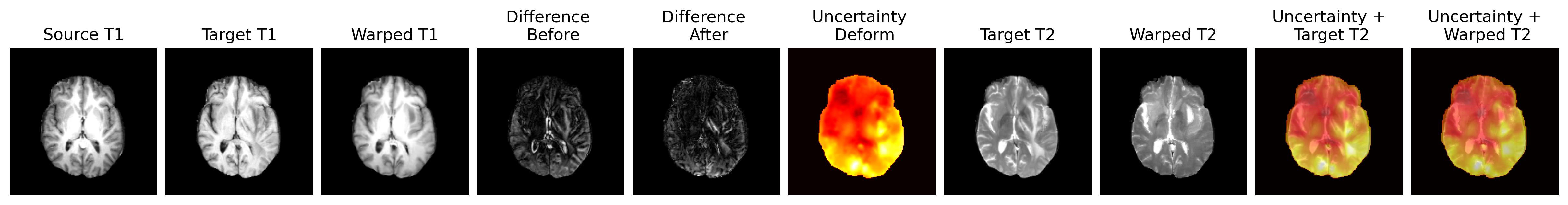}
    \caption{Case 034.}
    \label{appendix:fig:exp_case_study_one_bratsreg_034}
\end{subfigure}
\begin{subfigure}[b]{\textwidth}
    \centering
    \includegraphics[width=\linewidth]{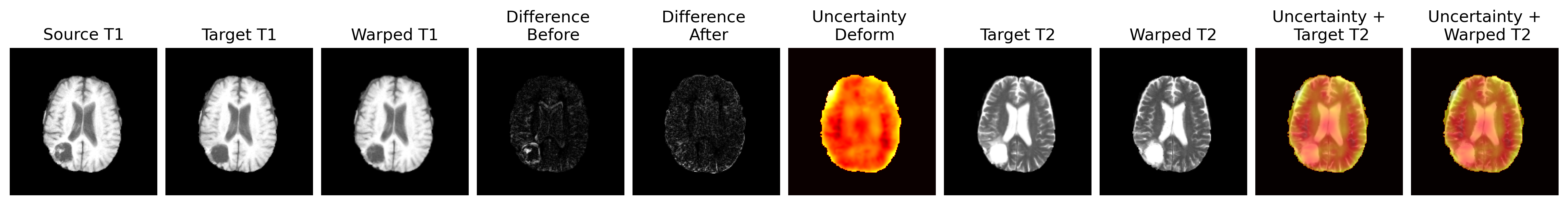}
    \caption{Case 037.}
    \label{appendix:fig:exp_case_study_one_bratsreg_037}
\end{subfigure}
\begin{subfigure}[b]{\textwidth}
    \centering
    \includegraphics[width=\linewidth]{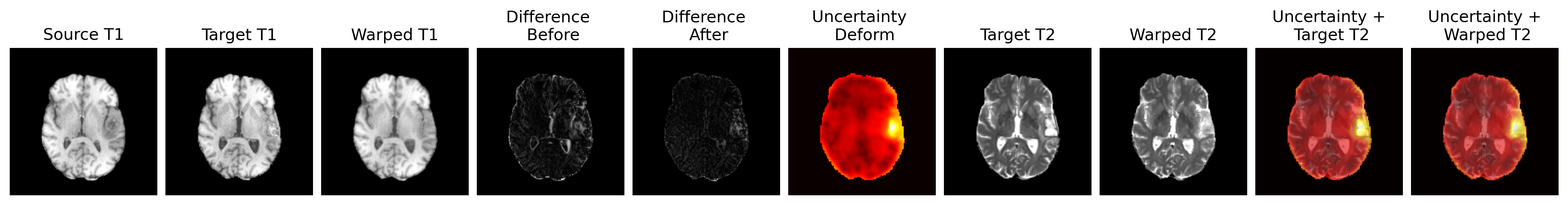}
    \caption{Case 080.}
    \label{appendix:fig:exp_case_study_one_bratsreg_080}
\end{subfigure}
\caption{Uncertainty map estimated during the registration between pre-operative and follow-up MRIs.}
\label{appendix:fig:exp_case_study_one_bratsreg}
\end{figure}

\clearpage

\subsubsection{Case Study of CBCT and CT with Different FOVs}

We present the results for three additional pairs of images from the Learn2Reg ThoraxCBCT dataset \citep{hugo2016data, hugo2017longitudinal} with the same experiment setting in \cref{sec:exp_case_study}. As shown in \cref{appendix:fig:exp_case_study_two_different_fov}, the CBCT/CT registrations consistently exhibit elevated uncertainty in the truncated regions above and below the lung, whereas the control pairs ($CT^\prime$/CT) exhibit uniformly low uncertainty. This pattern is observed across all three tested pairs, reinforcing the conclusion drawn in \cref{sec:exp_case_study}: the proposed uncertainty map reliably highlights regions of anatomical inconsistency caused by differences in FOV, and the finding generalizes beyond the specific examples shown in the main manuscript.

\begin{figure}[ht]
    \centering
    \begin{subfigure}[b]{\textwidth}
        \centering
        \includegraphics[width=\linewidth]{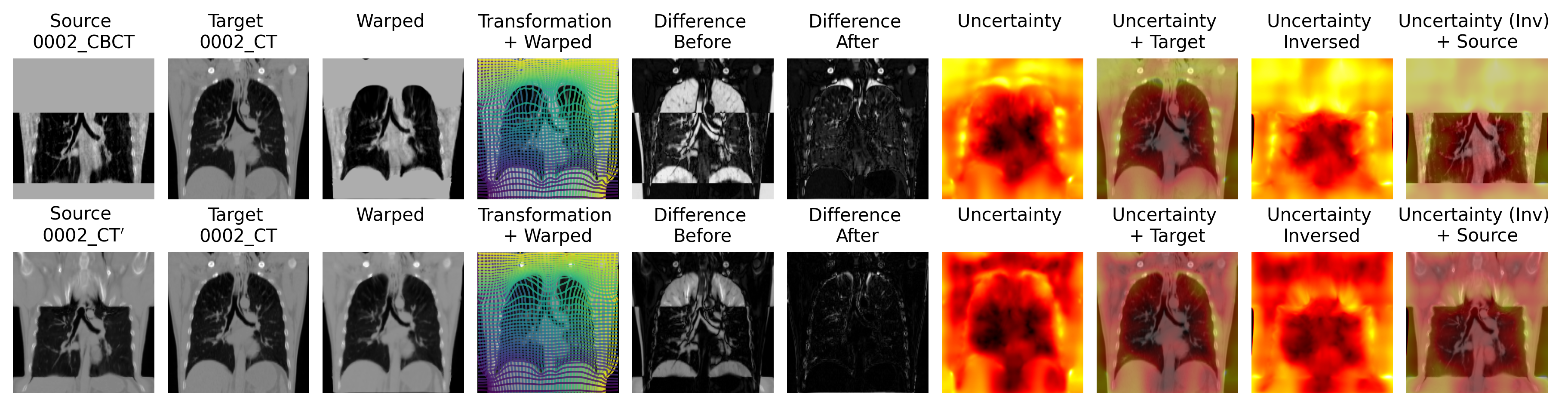}
        \caption{Case 0002.}
        \label{appendix:fig:exp_case_study_two_different_fov_0002}
    \end{subfigure}
    \begin{subfigure}[b]{\textwidth}
        \centering
        \includegraphics[width=\linewidth]{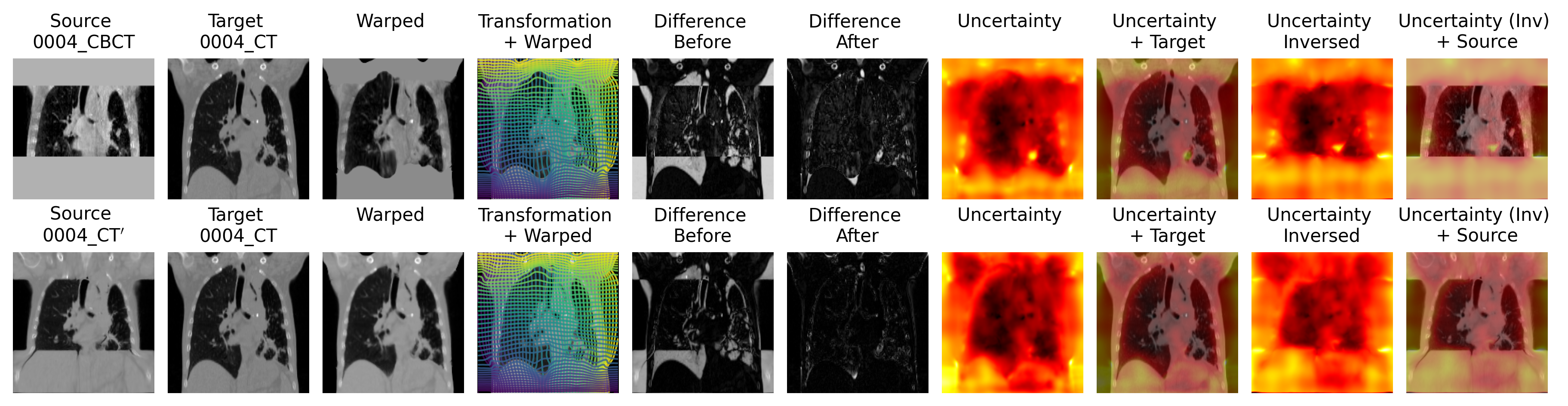}
        \caption{Case 0004.}
        \label{appendix:fig:exp_case_study_two_different_fov_0004}
    \end{subfigure}
    \begin{subfigure}[b]{\textwidth}
        \centering
        \includegraphics[width=\linewidth]{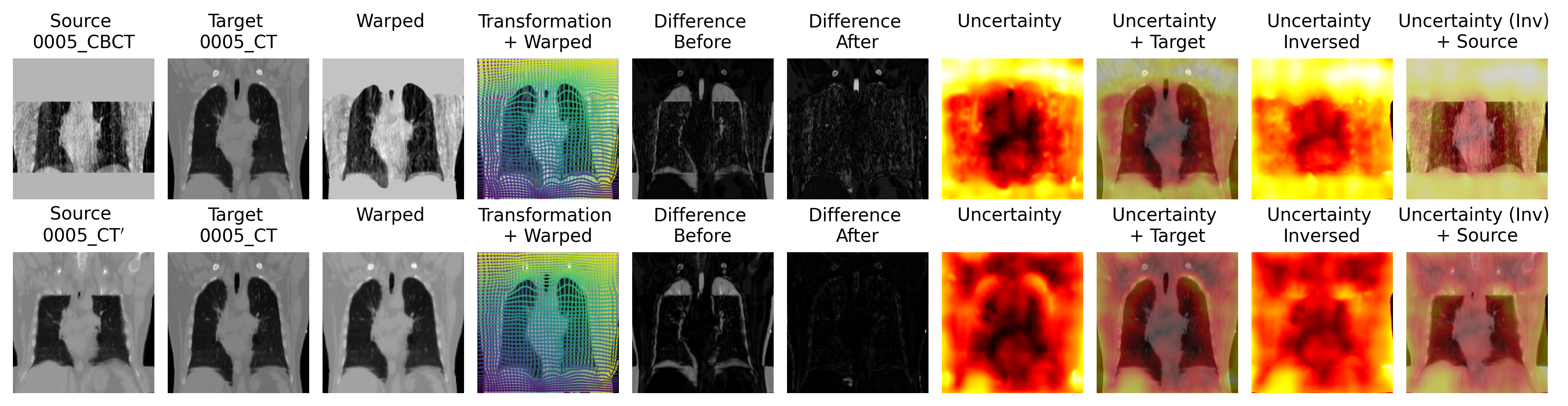}
        \caption{Case 0005.}
        \label{appendix:fig:exp_case_study_two_different_fov_0005}
    \end{subfigure}
    \caption{Uncertainty map estimated during the registration between CBCT and CT, which have different fields of view.  }
    \label{appendix:fig:exp_case_study_two_different_fov}
\end{figure}

\end{document}